\documentclass[runningheads]{llncs}
 
\usepackage{eccv}

\usepackage{eccvabbrv}

\usepackage{graphicx}
\usepackage{booktabs}
\usepackage{multirow}
\usepackage{siunitx}
\usepackage[table]{xcolor}  
\usepackage{colortbl}       
\usepackage{wrapfig}
\definecolor{lightgray}{gray}{0.93}
\DeclareRobustCommand{\circled}[1]{%
  \tikz[baseline=(char.base)]{
    \node[shape=circle,draw,inner sep=1pt, minimum size=1.1em] (char)
    {\normalfont\small #1};
  }%
}
\newcommand{\github}{\raisebox{-1.5pt}{\includegraphics[height=1.05em]{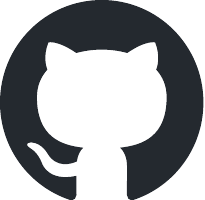}}\xspace}
\newcommand{\huggingface}{\raisebox{-1.5pt}{\includegraphics[height=1.01em]{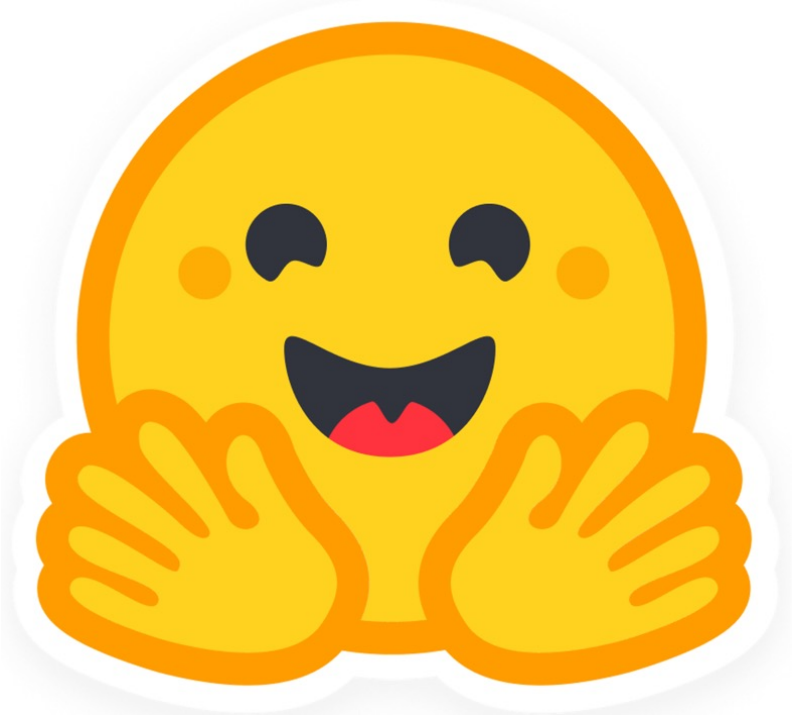}}\xspace}

\usepackage[accsupp]{axessibility}  
\usepackage{threeparttable} 

\usepackage[pagebackref,colorlinks=true,citecolor=blue]{hyperref}

\usepackage{orcidlink}

\usepackage{makecell}

\usepackage{caption}
\usepackage{tabularx}
\usepackage{pifont}
\usepackage{enumitem}

\usepackage{multirow}
\usepackage{inconsolata}
\usepackage[table,xcdraw]{xcolor}
\usepackage[most]{tcolorbox}

\usepackage{tabularx}
\usepackage{booktabs}
\usepackage[export]{adjustbox} 

\newcommand{\cmark}{\textcolor{Green}{\ding{51}}}%
\newcommand{\xmark}{\textcolor{Red}{\ding{55}}}%

\newcommand{\dataset}{\texttt{CRAG-\allowbreak MM-\allowbreak Diagnostics}\xspace}

\usepackage{xcolor}

\begin{document}

\title{\texttt{CRAG-MM-Diagnostics}: Enabling Stage-Wise Analysis of Knowledge-Intensive VQA}

\titlerunning{\texttt{CRAG-MM-Diagnostics}}

\author{
Hanseok Oh\inst{1,3}\thanks{Work conducted during an internship at Mila.} \orcidlink{0000-0003-2639-7117} \and
Parishad BehnamGhader\inst{2,3}\orcidlink{0009-0001-6117-8764} \and
Benno Krojer\inst{2,3}\orcidlink{0000-0003-1649-428X}, \\
Hyunji Lee\inst{4}\orcidlink{0000-0002-0187-336X} \and
Paul Liang\inst{5}\orcidlink{0000-0001-7768-3610} \and
Siva Reddy\inst{2,3,6}\orcidlink{0000-0003-3753-0323} \and
Verna Dankers\inst{2,3}\orcidlink{0000-0001-5955-6896}
}

\authorrunning{H. Oh et al.}


\institute{
    New York University, New York NY, USA \and
    McGill University, Montreal, Canada \and
    Mila - Quebec AI Institute, Montreal, Canada \and
    UNC Chapel Hill, Chapel Hill NC, USA \and
    MIT, Cambridge MA, USA \and
    Canada CIFAR AI Chair\\
\email{{hanseok.oh@nyu.edu}} \vspace{.5cm} \\ 
    \github \href{https://github.com/McGill-NLP/crag-mm-diagnostics}{McGill-NLP/crag-mm-diagnostics}\hspace{5mm}
    \huggingface \href{https://huggingface.co/collections/McGill-NLP/crag-mm-diagnostics}{CRAG-MM-Diagnostics} \vspace{-0.5cm}
} 

\maketitle

\begin{abstract}
  \textit{Knowledge-Intensive Visual Question Answering} (KI-VQA) benchmarks evaluate \textit{Vision–Language Models} (VLMs) as multimodal knowledge assistants by requiring external information beyond a provided image to answer questions. KI-VQA involves multiple sub-problems ---referring expression understanding, visual grounding, object recognition, knowledge retrieval, and reasoning---yet existing benchmarks typically report only end-task accuracy, obscuring where failures arise.
To analyze the \textit{full} KI-VQA pipeline, we introduce \dataset, a diagnostic benchmark with stage-wise data annotations that isolate \circled{1} language-based visual grounding, \circled{2} object identification, and \circled{3} knowledge retrieval and reasoning. We evaluate fully parametric and retrieval-augmented VLMs, providing fine-grained analyses using newly collected metadata, such as target ROIs, entity names, and visual complexity scores.
Our results point to knowledge retrieval and reasoning as the primary bottleneck, but also highlight issues in the other parts of the KI-VQA pipeline, such as the fact that VLMs struggle with target object identification or that image retrievers struggle to integrate textual cues.
These findings expose fundamental limitations in current KI-VQA systems and motivate stage-aware evaluation. We, lastly, leverage these findings to propose a grounded bimodal RAG pipeline that integrates a visual grounding module to crop targets before image retrieval, boosting \texttt{GPT-5} and \texttt{Qwen}'s respective accuracies by 13.3 and 8.5 percentage points.

\end{abstract}

\section{Introduction}

\begin{figure}[t]
    \centering
    \includegraphics[width=\linewidth]{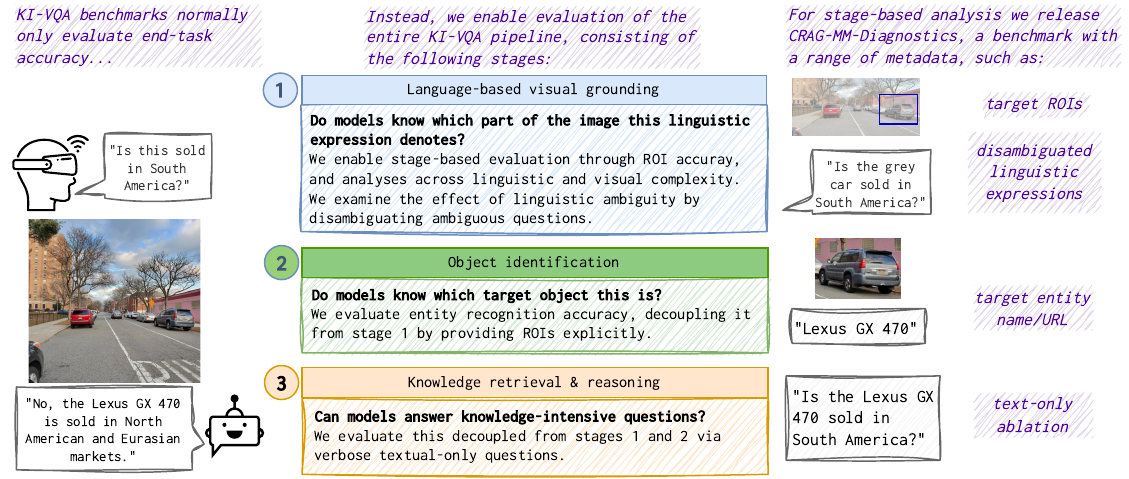}
    \caption{
    Illustrative summary of the stage-wise evaluation enabled by \dataset, thanks to new types of metadata we add. The three stages evaluated are (1) language-based visual grounding, (2) object identification, and (3) knowledge extraction \& reasoning, and are further detailed in the figure.
    }
    \label{fig:pipeline_illustration}
\end{figure}

Recent advances in \textit{Vision-Language Models}~(VLMs) have accelerated interest in their deployment in real-world applications~\cite{cragMM2025, Narayan2025DeepMMSearchR1EM, deitke2025molmo, Bai2025Qwen25VLTR}. 
A core requirement for these assistants is the ability to interpret visual scenes and provide informative responses, typically evaluated through \textit{Visual Question Answering}~(VQA) benchmarks. However, while standard VQA often focuses on describing visible attributes, there is growing attention for \textit{Knowledge-Intensive} VQA~(KI-VQA). In this setting, the question cannot be answered solely from the image but also requires external world knowledge~\cite{ijcai2017p179, marino2019ok, chen2023can, Mensink2023EncyclopedicVV,kabir2024comprehensive}.
KI-VQA is a vital step towards a future in which VQA is seamlessly integrated in daily life, in which one can ask complex questions on the go, when walking down the street using smart glasses, or texting photos back and forth with a friend. 

However, existing KI-VQA benchmarks often fail to capture real-world visual complexity, instead relying on images with salient, centered objects and minimal clutter~\cite{marino2019ok,Mensink2023EncyclopedicVV,chen2023can}. 
In contrast, real-world scenes contain multiple, often visually similar objects, with less salient and off-center targets (see \Cref{fig:pipeline_illustration}).
Challenging benchmarks that do include visually complex images often focus only on end-task evaluation~\cite{hu2025mragbench,cragMM2025}, obscuring the root causes of failure (i.e., whether errors stem from flawed visual grounding, incorrect entity identification, or downstream reasoning). 
For example, when a model fails to answer the question in \Cref{fig:pipeline_illustration}, is the issue target identification or missing knowledge?

In this work, we take a step towards a more fine-grained analysis of KI-VQA, aiming to understand how models perform in intermediate stages of the pipeline.
Unlike the majority of KI-VQA studies that only evaluate end-task accuracy, we focus on three intermediate stages: 
\circled{1} \textbf{language-based visual grounding}, used to localize the target region of interest in a complex image~\cite{wu2024v,qiang2025ver}; \circled{2} \textbf{object identification}, used to map the localized region to a canonical entity (e.g., a Wikipedia page)~\cite{Mensink2023EncyclopedicVV,Narayan2025DeepMMSearchR1EM}; and \circled{3} \textbf{knowledge retrieval and reasoning}, which answers the question after possible textual and visual information retrieval~\cite{marino2019ok, chen2023can}.
To enable this analysis, we introduce \dataset, a benchmark built on top of \texttt{CRAG-MM} \cite{cragMM2025}, as described in \Cref{sec:dataset}.
While \texttt{CRAG-MM} provides a foundation for KI-VQA in egocentric, visually complex scenes, it serves primarily as an end-to-end benchmark without the structural markers needed to diagnose internal model failures. \dataset transforms this resource into a diagnostic framework by introducing stage-specific annotations.
\Cref{fig:pipeline_illustration} illustrates the three stages, and a small subset of the new metadata.

Using the extended benchmark, we conduct a comprehensive analysis for the three stages in Sections~\ref{sec:grounding},~\ref{sec:identification} and~\ref{sec:rag}, using fully parametric VLMs, models specialized for each stage, and retrieval-augmented models.
Our analysis uncovers several fundamental limitations, among which:
\begin{itemize}
    \item Visual grounding of target objects is more challenging for proprietary models;
    \item Linguistic ambiguity in QA pairs affects the entire KI-VQA pipeline;
    \item VLMs, in particular compared to specialized models, struggle to identify target objects that are unpopular;
    \item Localizing the target object improves image retrieval substantially, especially in ambiguous queries;
    \item Most KI-VQA errors are due to inadequate knowledge retrieval and reasoning rather than incorrect object identification (e.g., only 20.9\% of \texttt{GPT-5}'s errors can be resolved by providing the ground-truth target object name).
\end{itemize}
By providing a stage-aware analysis, we offer actionable insights for the design of robust KI-VQA systems. In \Cref{sec:rag}, we leverage our findings to study a modeling pipeline combining components from all stages to achieve better performance, boosting \texttt{GPT-5} and \texttt{Qwen}'s respective accuracies by 13.3 and 8.5 percentage points.
We conclude in \Cref{sec:conclusion}.

\section{Related Work}

In this section, we discuss the evolution of KI-VQA and recent advances in diagnostic evaluation for VLMs. We provide a more detailed comparison in Appendix~\ref{app:dataset}.

\subsection{Knowledge-Intensive Visual Question Answering Benchmarks} 

Since the introduction of VQA \cite{antol2015vqa}, later benchmarks have increased difficulty by requiring external knowledge, leading to the development of \textit{knowledge-intensive} VQA (KI-VQA). 
\textit{OK-VQA} \cite{marino2019ok} focuses on commonsense and simple factual knowledge. 
Subsequent datasets impose more specialized knowledge demands: \textit{EncyclopedicVQA} \cite{Mensink2023EncyclopedicVV} and \textit{SnapNTell} \cite{qiu2024snapntell} target long-tail entities (e.g., ``\textit{Pinus pinea}''), while \textit{ViQuAE} \cite{Lerner2022ViQuAEAD} and \textit{InfoSeek} \cite{chen2023can} emphasize named entities and realistic information-seeking scenarios. These benchmarks typically require retrieving external knowledge.

Despite this progress, most KI-VQA benchmarks increase difficulty primarily along the \emph{language axis}—e.g., by requiring complex reasoning \cite{tran2025reasonvqa} or including temporally evolving knowledge \cite{li2025benchmarking}—while implicitly assuming object recognition is trivial or based on salient cues \cite{chen2023can,Mensink2023EncyclopedicVV,qiu2024snapntell}. 
In contrast, fine-grained visual benchmarks focus on challenges along the \textit{visual axis}. Datasets such as \textit{V*} \cite{wu2024v}, \textit{GigaGrounding} \cite{Ma2024WhenVG}, and \textit{MME-RealWorld} \cite{zhangmme} focus on visually challenging scenarios, including crowded, high-resolution, or remote-sensing images where targets are small or non-salient, requiring precise localization. However, these benchmarks are mainly studied outside information-seeking settings.

Such visual difficulty becomes even more critical when retrieval is involved, as external tools must correctly identify subtle targets under noisy conditions—an aspect largely overlooked in KI-VQA.
Recent multimodal \textit{Retrieval Augmented Generation} (RAG) benchmarks, such as \textit{MRAG-Bench} \cite{hu2025mragbench} and \textit{CRAG-MM} \cite{cragMM2025}, incorporate realistic visual challenges like viewpoint shifts, occlusion, and egocentric perspectives.
However, these benchmarks lack fine-grained diagnostics to attribute failures to specific visual factors or to disentangle visual and linguistic errors.
We, therefore, present \dataset, which provides fine-grained stage-wise diagnostics for \emph{language-based visual grounding}, \emph{object identification}, and \emph{knowledge retrieval and reasoning}.

\subsection{Diagnostic evaluation of VLMs}
Recent studies have proposed diagnostic evaluations of intermediate stages to better understand VLMs' capabilities in multimodal tasks. 
\textit{MMVet} \cite{yu2024mm} annotates examples from 16 multimodal tasks with six core vision–language skills, which are not specifically curated for the KI-VQA task.
\textit{HallusionBench} \cite{guan2024hallusionbench} decomposes hallucinations into language hallucination and visual illusion to analyze failure modes.
Similarly, \textit{Prism} \cite{qiao2024prism} disentangles perception from reasoning in general VQA.
Despite these advances, diagnostic evaluation for \emph{knowledge-intensive} information seeking tasks remains limited. 
Existing evaluations are often confined to isolated visual recognition tasks \cite{Hu2023OpendomainVE} or rely solely on final QA accuracy \cite{marino2019ok,chen2023can, Mensink2023EncyclopedicVV}, overlooking realistic KI-VQA scenarios that require precise grounding and contextual target identification \cite{xu2025mc,ma2025deepperception}.
The recent KI-VQA benchmark \textit{VisualSimpleQA} \cite{Wang2025VisualSimpleQAAB} partially addresses this gap by comparing multimodal and text-only questions, but still focuses on end-task accuracy and is limited to parametric VLMs. 
\section{The curation of \dataset}\label{sec:dataset}
To facilitate fine-grained KI-VQA analysis, we develop a diagnostic benchmark (\texttt{CRAG-MM-Diagnostics}) by augmenting and refining \texttt{CRAG-MM} \cite{cragMM2025}.
\texttt{CRAG-MM} consists of diverse (image, question, answer) triplets spanning 13 domains, including nearly 1.9K single-turn egocentric images that mimic captures from wearable devices. This egocentric setting enables \textit{realistic} KI-VQA evaluation and provides sufficient visual diversity and complexity to analyze performance across different levels of visual saliency---e.g., by varying object sizes or scene crowdedness.
However, it only permits end-to-end QA analysis, making it impossible to investigate the ability of different KI-VQA components separately.

\paragraph{Human annotations.}
To ensure the benchmark's quality, we apply an initial filtering process prior to human annotation, removing samples that are not \textit{knowledge-intensive} or concern dynamically changing knowledge, as detailed in Appendix~\ref{app:dataset-stats}.
We then manually annotate textual metadata and target object regions using bounding boxes.\footnote{Some metadata annotations were model-assisted via preliminary annotations, as detailed in Appendix~\ref{app:dataset_metadata}.}
Each sample in the dataset is annotated once by an author and consecutively double-checked by another.
The metadata added through human annotation, illustrated using \Cref{fig:pipeline_illustration}, includes:

\begin{enumerate}[topsep=0pt]
    \item \textbf{Entity name, Wikipedia URL}: the target object's name and corresponding Wikipedia URL (e.g., ``Lexus GX 470'' in case of \Cref{fig:pipeline_illustration}).
    \item \textbf{Referring expression type}: the queries' labels that distinguish \textit{salient} cases (e.g., where ``this van'' suffices as a referral to identify a unique object) from \textit{ambiguous} ones
    (e.g., the non-specific expression ``this'' in \Cref{fig:pipeline_illustration}),
    and ones where there is an \textit{in-image cue} (e.g., ``the red car'') from a \textit{knowledge-intensive cue} (e.g., recognizing the service associated with a logo to identify that ``this restaurant'' refers to a specific building). 
    See \Cref{tab:example} for examples.
    \item \textbf{Disambiguated question}: for examples with ambiguous referring expression, we add a disambiguated question. For instance, for \Cref{fig:pipeline_illustration}, this question replaces ``this car'' with ``the grey car on the right''.
    \item \textbf{Text-only question}: for all examples, we add a question which allows models to rely on the textual modality only, by inserting the correct entity into the question, e.g., ``Is the Lexus GX 470 sold in South America?'' for \Cref{fig:pipeline_illustration}.
    \item \textbf{Target region}: in addition to the textual metadata mentioned above, the human annotation also involves the selection of a bounding box in which the target object is contained, as previously demonstrated in \Cref{fig:pipeline_illustration}.
\end{enumerate}

After annotation, we removed 95 samples due to missing Wikipedia links for the target entity, mismatched image-QA pairs, questions that do not require visual information, low-quality images that hinder localization (e.g., due to blur), and time-dependent questions (e.g., “last year”) identified during the annotation.

\paragraph{Automated annotations.}
Additionally, we automatically augment the data with metadata capturing an object's popularity, and its visual complexity: 
\begin{enumerate}[topsep=0pt]
    \item[6.] \textbf{Target popularity}: the Wikipedia page views in 2025, following ~\cite{Mallen2022WhenNT,qiu2024snapntell}.
    \item[7.] \textbf{Target object size}: measured as the ratio of the target’s bounding box area to the image area.
    \item[8.] \textbf{Distance to center}: computed as the distance between the target centroid and the image center.
    \item[9.] \textbf{Scene crowdedness}: approximated by the number of objects detected using the open-set detector Grounding-DINO~\cite{Liu2023GroundingDM}.
\end{enumerate}

\noindent Using the three visual complexity dimensions (metadata 7–9), we define a summary metric, \textbf{visual saliency}, to quantify a target’s prominence in an image (see Appendix~\ref{app:dataset_metadata}).
After filtering and annotation, \texttt{CRAG-MM-Diagnostics} contains 1,149 samples. Dataset statistics are shown in \Cref{fig:dataset-stats}; example QA pairs, further metadata details, the annotation protocol and the annotation tool are described in Appendix~\ref{app:dataset}.
\section{Stage \circled{1} Language-based Visual Grounding} \label{sec:grounding}

We use \texttt{CRAG-MM-Diagnostics} to systematically evaluate the KI-VQA stages (as shown in \Cref{fig:pipeline_illustration}), beginning with the system's ability to localize and identify the subject of a referring expression within a visual scene. We assess this \textbf{language-based visual grounding} ability by measuring how accurately models predict the bounding box of a target object across varying linguistic and visual variations. We first describe the experimental setup. 

\subsection{Experimental Preliminaries}
\label{sec:grounding_models}

\paragraph{Models.} 
Our evaluation spans both generalized VLMs (which are capable of end-to-end KI-VQA) and specialized models designed specifically for this step of the pipeline (i.e., localization). 
We examine the performance of both, to inform building KI-VQA solutions that integrate specialized models, which we will perform later on, in \Cref{sec:rag} (when performing stage \circled{3} evaluation).

Among the \textbf{generalized models} that jointly encode images and text, we include open-source architectures \texttt{Llama-3.2-11B} \cite{meta2024llama} and the \texttt{Qwen2.5-VL} family  (ranging from 3B to 72B parameters) \cite{Bai2025Qwen25VLTR}, alongside proprietary state-of-the-art models \texttt{GPT-5} and \texttt{GPT-5-mini}. 
The \textbf{specialized models} focus explicitly on alignment between language and visual regions: \texttt{Grounding-DINO} \cite{Liu2023GroundingDM} and \texttt{OWL-ViT} \cite{minderer2022simple} perform open-vocabulary object detection by aligning text embeddings with image features for precise localization.
Detailed prompts and details for this task are included in Appendix~\ref{app:visual_grounding}; the prompts provide models with the question and image, and require outputting precise bounding box coordinates.

\paragraph{Evaluation metric.}
Following established related work on visual grounding evaluation \cite{xiao2025towards, Ma2024WhenVG}, we measure \textit{Intersection over Union} (IoU) between the predicted and ground-truth bounding boxes of the target object.
A prediction is classified as correct if the IoU exceeds a threshold of 0.5.

\begin{figure}[t]
    \centering
    \includegraphics[trim={0 1cm 0 0},width=0.9\linewidth]{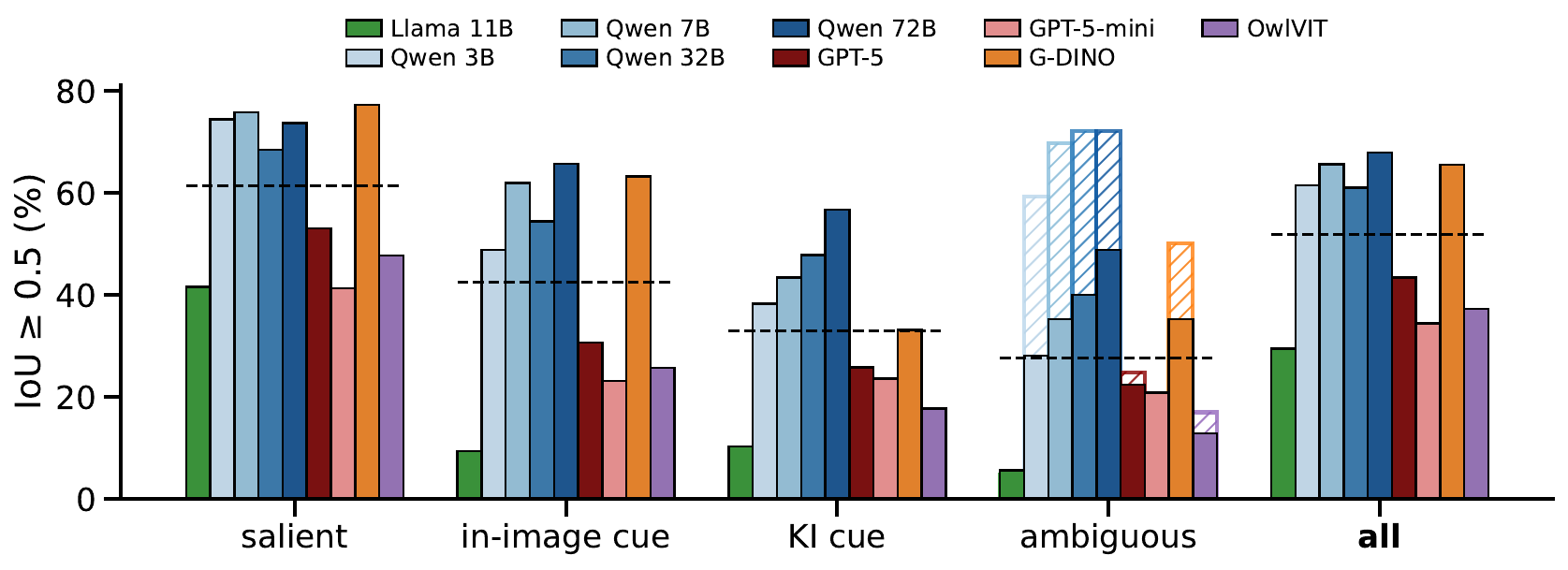}
    \caption{
    Grounding accuracy per referring expression type (horizontal lines: mean; hatched: disambiguated query).
    It shows that: (1) Drops on non-salient and ambiguous targets reveal high sensitivity to expression type. (2) Gains from disambiguation suggest many failures arise from query ambiguity rather than visual processing deficits.
    }
    \label{fig:grounding_accuracies}
\end{figure}

\subsection{Analyses and Findings}
\Cref{fig:grounding_accuracies} details the IoU per model, per referring expression type, and across all examples through the `\textbf{all}' bar.
The \texttt{Qwen2.5-VL} family demonstrates robust grounding across all sizes, peaking at 67.8\% accuracy.
\texttt{Llama-3.2} and the \texttt{GPT-5} variants, however, have markedly lower performance.
The fact that even the smallest \texttt{Qwen} is quite good at this task is likely because the model family was explicitly trained to understand bounding boxes \cite{Bai2025Qwen25VLTR}; the \texttt{GPT-5} results, in particular, underscore that spatial understanding of images does not naturally emerge, even in SOTA models.
Even if \texttt{GPT-5} \textit{can} perform KI-VQA in general, the fact that it struggles with explicit grounding in this first stage limits the explainability and interpretability of the model.
When contrasting the generalized and specialized models, it is noteworthy that \texttt{Grounding-DINO} performs remarkably well; despite its modest 0.2B parameters, it rivals the \texttt{Qwen} models in accuracy, presenting a highly efficient alternative for modular KI-VQA pipelines that include grounding.

\begin{figure}[t]
    \centering
    \includegraphics[width=0.65\linewidth]{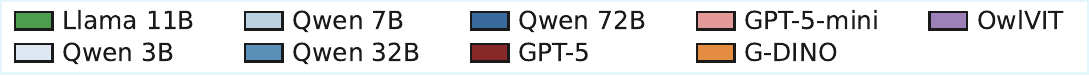}\\
    \begin{subfigure}[b]{0.24\textwidth}
        \centering
        \includegraphics[width=\textwidth]{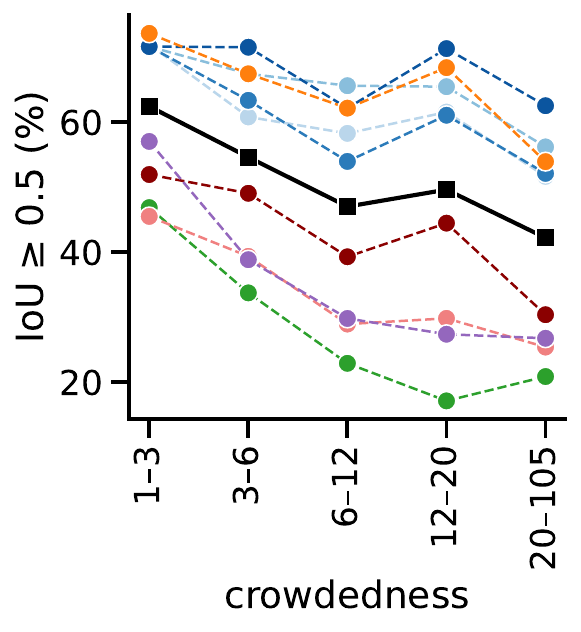}
    \end{subfigure}
    \hfill
    \begin{subfigure}[b]{0.24\textwidth}
        \centering
        \includegraphics[width=\textwidth]{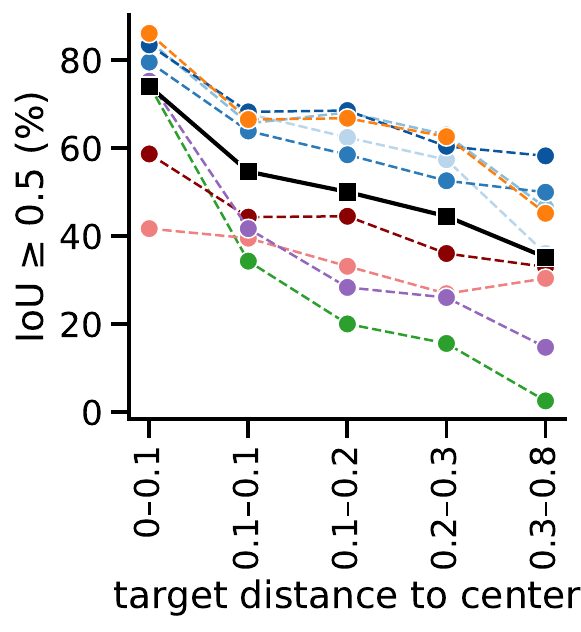}
    \end{subfigure}
    \hfill
    \begin{subfigure}[b]{0.24\textwidth}
        \centering
        \includegraphics[width=\textwidth]{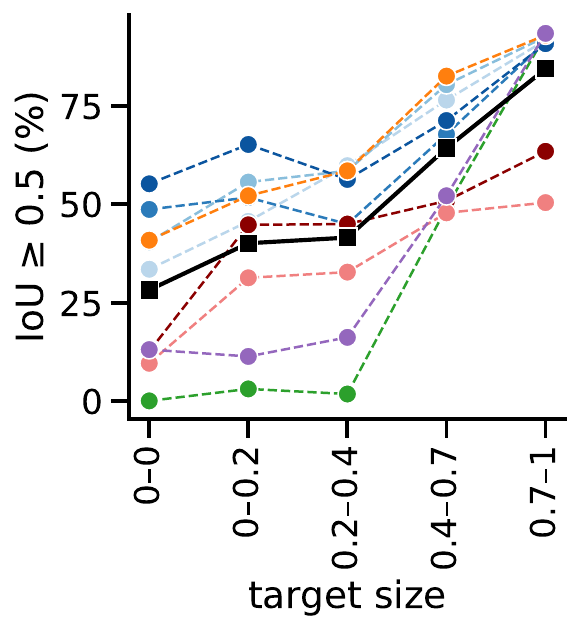}
    \end{subfigure}
    \hfill
    \begin{subfigure}[b]{0.24\textwidth}
        \centering
        \includegraphics[width=\textwidth]{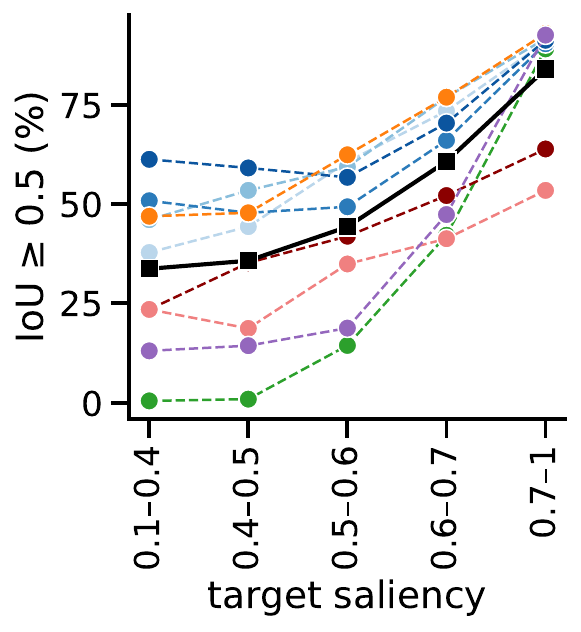}
    \end{subfigure}
    \caption{Visual grounding scores along four dimensions of visual complexity (black squares indicate model mean). These plots show that low target saliency is the primary bottleneck, acutely compounding performance drops caused by scene density, off-center placement, and small target size.
    }
    \label{fig:grounding_per_visual_complexity}
\end{figure}

\paragraph{Referring expressions highly influence performance.}

Overall performance obscures the impact of the \textit{type} of referring expression.
The `referring expression' metadata we include in \dataset distinguishes salient cases from more complex ones, which constitute 36.6\% of the dataset. Appendix~\ref{app:dataset_examples} provides concrete examples of the different types of expressions.
\Cref{fig:grounding_accuracies} shows that 
while salient targets are the most straightforward to ground, ambiguous expressions represent a severe challenge.
This gap is especially pronounced for smaller models (e.g., \texttt{Qwen} 3B and 7B), highlighting the role of model capacity in handling linguistic nuance and multimodal integration. Moreover, results demonstrate that although \texttt{Grounding-DINO} performs similarly to the \texttt{Qwen} family overall, it falls short of the larger models in samples with complex referring expressions (i.e., in the case of a `knowledge-intensive cue' and `ambiguous' examples).

\paragraph{Ambiguity is a bottleneck for visual grounding.}
To isolate the effects of ambiguity, we compared performance on samples with ambiguous queries against disambiguated ones (e.g., changing ``this'' to ``the grey car on the right'' in \Cref{fig:pipeline_illustration}).
Disambiguation resulted in a substantial performance surge, a mean 44.0\% improvement (see hatched bars in \Cref{fig:grounding_accuracies}). This indicates that a lack of linguistic specificity is a primary bottleneck.
The gains are largest for \texttt{Qwen} models, whereas disambiguated queries do not improve performance for \texttt{Llama} and \texttt{GPT-5-mini}, underscoring that their subpar performance is a more fundamental issue with these models' spatial reasoning abilities rather than merely the result of ambiguity.
The presence of ambiguous queries in \texttt{CRAG-MM} is likely due to the data's egocentric nature. If a QA pair does not capture the user's gaze or intent, evaluation using the original queries may underestimate models' true KI-VQA performance.

\paragraph{Grounding degrades as visual complexity grows.}
Language-based visual grounding depends not only on the referring expression but also on image complexity and target saliency \cite{wu2024v, Ma2024WhenVG}. \Cref{fig:grounding_per_visual_complexity} demonstrates this for the four visual complexity axes previously introduced in \Cref{sec:dataset}.
Performance decreases with increased crowdedness and distance---with \textit{point biserial} (pb) correlations of $r_{pb}=-.117$ and $-.254$, respectively, averaged over models---and improves with larger size and higher saliency---$r_{pb}=.422$ and $.391$. 

\begin{tcolorbox}[colback=Cerulean!10!white, colframe=Cerulean!75!black,left=1mm,right=1mm]
Intermediate takeaways based on visual grounding (stage \circled{1}) evaluation:
\begin{enumerate}[topsep=0pt]
    \item VLMs lack spatial awareness without being explicitly trained for this, as \texttt{Llama} and \texttt{GPT-5(-mini)} struggle a lot more than \texttt{Qwen} when predicting bounding boxes;
    \item Complex referring expressions and ambiguity negatively affect visual grounding;
    \item In line with prior work~\cite{wu2024v,Ma2024WhenVG} various factors of visual complexity degrade grounding performance;
    \item \texttt{Grounding-DINO} best balances performance with efficiency.
\end{enumerate}
\end{tcolorbox}

\section{Stage \circled{2} Object Identification} \label{sec:identification}

Next, we study the models' ability to perform \textbf{target object identification}; a vital step towards KI-VQA since the vast majority of \dataset questions concern specific named entities. Here, we first elaborate on the experimental setup prior to diving into our findings. We evaluate models by providing the original image and question, alongside task-specific instructions (detailed in Appendix~\ref{app:object_identification}), and require the models to output the name of the target entity.

\subsection{Experimental Preliminaries}

\paragraph{Models.}
Consistent with our previous analysis, we evaluate two model classes. Firstly, we consider \textbf{generalized models} (see \Cref{sec:grounding_models}), which rely solely on internal parametric knowledge. Evaluating them on object identification, therefore, isolates a step that these models normally perform implicitly during KI-VQA.
In contrast, \textbf{specialized models} retrieve images and metadata from external knowledge sources. They implicitly perform object identification when at least one retrieved item corresponds to the target.
Although this does not yield a direct comparison with generalized models, it allows us to examine complementary strengths of parametric vs.\ retrieval-based identification.
The specialized models are the unimodal retriever \texttt{CLIP-\allowbreak ViT-\allowbreak Large-\allowbreak Patch14-\allowbreak 336} \cite{Radford2021LearningTV} and the multimodal retrievers \texttt{VLM2Vec-V2.0} \cite{Meng2025VLM2VecV2AM} and \texttt{Qwen-3-VL-Embedding-2B} \cite{li2026qwen3}. We use the \texttt{CRAG-MM} image knowledge graph \cite{cragMM2025} as the retrieval corpus and recompute embeddings for each model.

\paragraph{Evaluation metric.}
For generalized models, we measure entity prediction accuracy using exact string match. For cases without an exact match, we utilize LLM-as-a-judge (\texttt{GPT-4o-mini}) to determine semantic equivalence. For specialized retrieval models, we report \texttt{recall@10}. 
Specifically, we perform normalized partial string matching between the ground-truth entity name and entity names associated with the top-10 retrieved images following related work \cite{singh2015implementation,ijcai2024p690}. 
We do not apply the LLM-as-a-judge approach here, as verifying semantic equivalence for all top-$k$ results from retrieval sources is very computationally expensive.

\subsection{Analyses and Findings}

\begin{figure}[t]
\begin{subfigure}[T]{0.25\columnwidth}
    \centering
    \includegraphics[height=3.5cm]{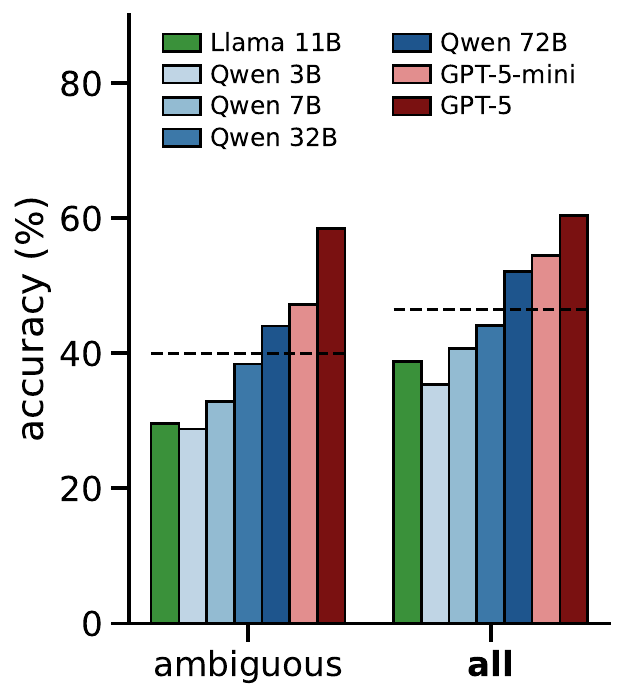}
    \caption{OI performance}
    \label{fig:oi_results_generalized}
\end{subfigure}
\begin{subfigure}[T]{0.45\columnwidth}
    \centering
    \includegraphics[height=3.5cm]{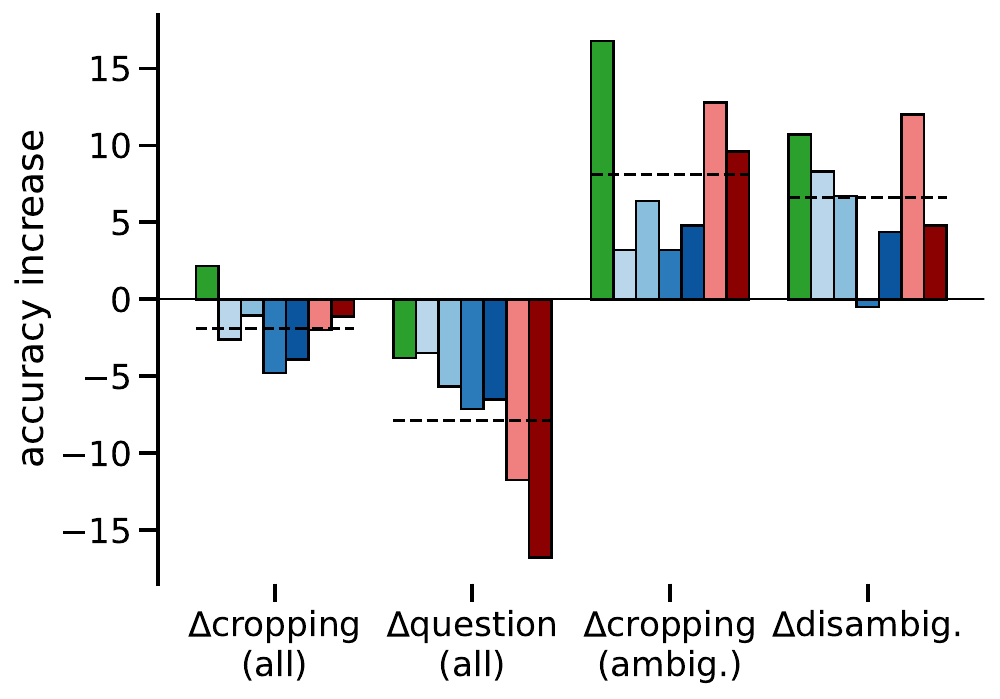}
    \caption{Effect of modifying the question/image}
    \label{fig:oi_results_transformations_generalized}
\end{subfigure}
\begin{subfigure}[T]{0.28\columnwidth}
    \includegraphics[height=3.5cm]{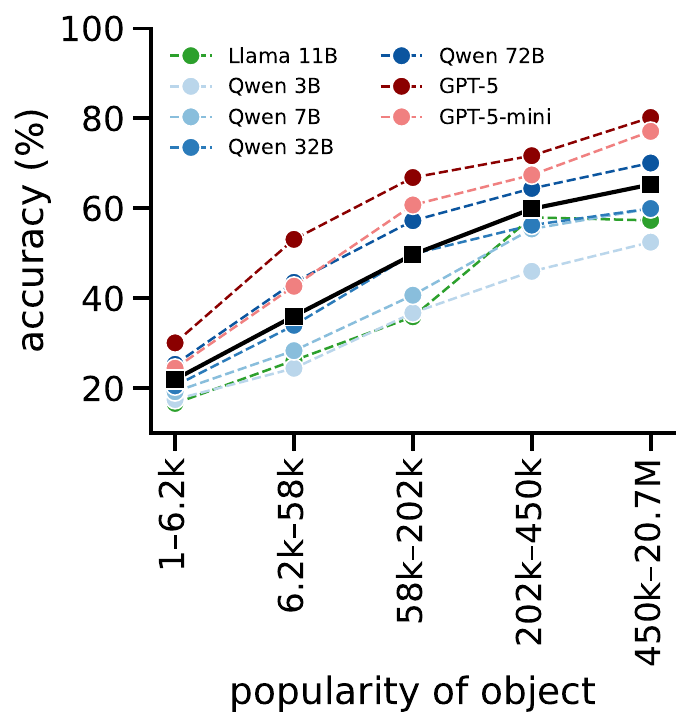}
    \caption{Accuracy vs popularity}
    \label{fig:popularity_generalized}
\end{subfigure}
\caption{
Performance breakdown of generalized models on the stage \circled{2} object identification task.
(b) Accuracy changes under input manipulations: $\Delta$ `cropping' marks the accuracy change after cropping the target object; $\Delta$ `question' marks the change after simplifying the question; $\Delta$ `disambig.'  marks the change after using the disambiguated questions for the ambiguous stimuli.
(c) Accuracy across target object popularity bins.  
}
\label{fig:object_identification_results_generalized}
\end{figure}

Before diving into fine-grained analyses, \Cref{fig:oi_results_generalized} reveals the overall performance and performance for examples with ambiguous referring expressions.
Although proprietary models struggled with visual grounding in \Cref{sec:grounding}, they now achieve the strongest object identification performance.
Grounding being a prerequisite for object identification suggests that while these models lack the specialized `language' of coordinate-based grounding, they possess a strong implicit grounding capability that facilitates identification.
Yet, even \texttt{GPT-5} struggles to predict the correct entity name for approximately 40\% of the examples, which will have implications for the end-task KI-VQA accuracy down the line.

\paragraph{VLMs benefit from multimodal cues.}
Next, we analyze the contribution of each modality to object identification for the generalized models (\Cref{fig:oi_results_transformations_generalized}):
\begin{itemize}
    \item[-] $\Delta$ cropping: 
    Providing a crop of the target region instead of the full image results in a slight net decrease in accuracy for most models, though it predictably aids ambiguous stimuli.
    This suggests that models might benefit from contextual cues beyond the target region for performing accurate object identification.
    \item[-] $\Delta$ question: 
    Replacing the original query with a simplified prompt (e.g., ``What is this object?'') reduces accuracy by 7.9 percentage points. This confirms that models leverage both visual and textual cues to refine their visual identification. An example of a useful textual cue is, for instance, if the question refers to an ``SUV'', greatly restricting the set of possible target objects.
    \item[-] $\Delta$ disambiguated: 
    Using disambiguated questions provides performance gains nearly on par with those of cropping, suggesting that ambiguous examples cause an underestimation of a model's latent KI-VQA capabilities.
\end{itemize}

\paragraph{VLMs are affected by objects' popularity.}
Next, we utilize other types of metadata to further understand what does or does not affect object identification performance when the generalized VLMs perform the task.
Unlike the grounding task in Stage \circled{1}, object identification is less sensitive to visual complexity;
in fact, saliency is (only weakly) negatively related to accuracy ($r_{pb}=-.168$) (described in Appendix~\ref{app:oi_generalized}).
A clearer trend, however, exists for the impact of popularity as depicted in \Cref{fig:popularity_generalized} (with an average Spearman's $\rho$ of 0.314): VLMs are better at identifying entities with high Wikipedia page-view counts, consistent with established literature on long-tail visual knowledge distribution~\cite{Hu2023OpendomainVE,Mensink2023EncyclopedicVV,qiu2024snapntell}.

\paragraph{Image retrievers benefit from visual isolation and struggle with linguistic comprehension.}
Specialized retrievers present a different performance profile.
\Cref{fig:oi_results_specialized} shows recall@10 for all examples versus ambiguous cases using image-only queries.\footnote{Detailed performance of different input variants and performance trends across different top-$k$ values are described in Appendix \ref{app:image_retriever}.}
These results confirm that object identification is a challenging task for retrievers, while \texttt{CLIP}'s performance stands out as relatively strong, when taking into account the size gap to the other models (427M vs 2B).
Notably, ambiguous cases suffer from poor performance even without textual queries, likely because these instances are more visually complex.\footnote{
Mean visual saliency is 0.38 for ambiguous cases vs. 0.58 for non-ambiguous ones.
} \Cref{fig:oi_results_transformations_specialized} demonstrates performance changes when varying the input to the specialized models:

\begin{figure}[t]
\begin{subfigure}[T]{0.23\columnwidth}
    \centering
    \includegraphics[height=3.5cm]{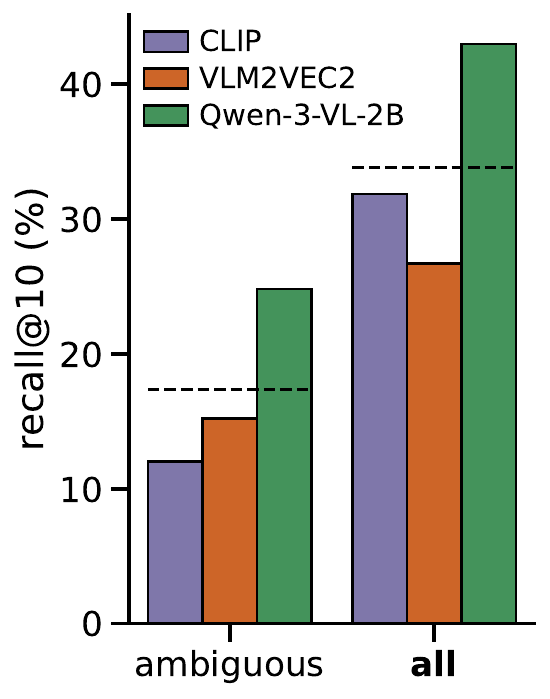}
    \caption{OI performance}
    \label{fig:oi_results_specialized}
\end{subfigure}
\begin{subfigure}[T]{0.43\columnwidth}
    \centering
    \includegraphics[height=3.5cm]{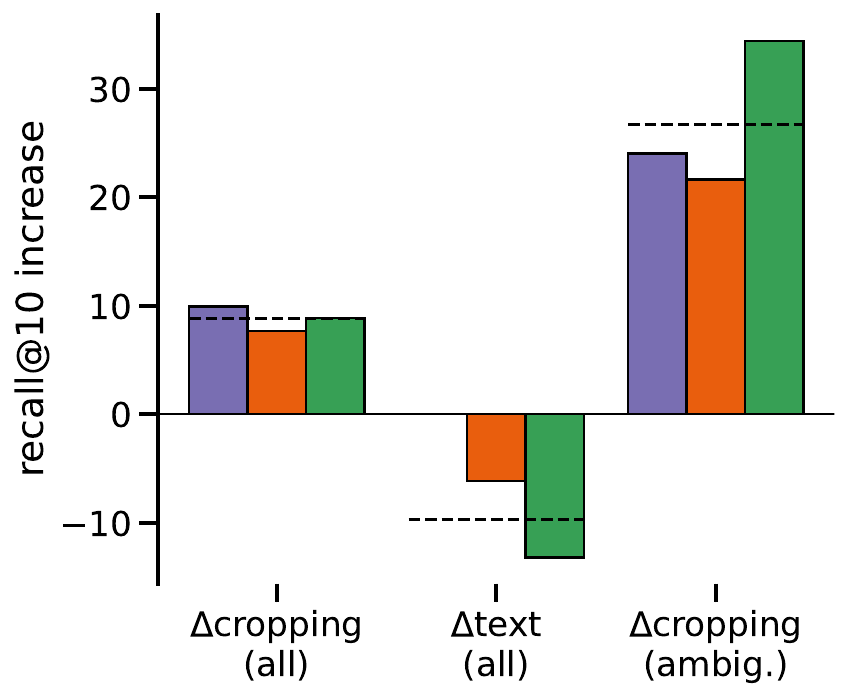}
    \caption{Effect of modifying the question/image}
    \label{fig:oi_results_transformations_specialized}
\end{subfigure}
\begin{subfigure}[T]{0.28\columnwidth}
    \includegraphics[height=3.5cm]{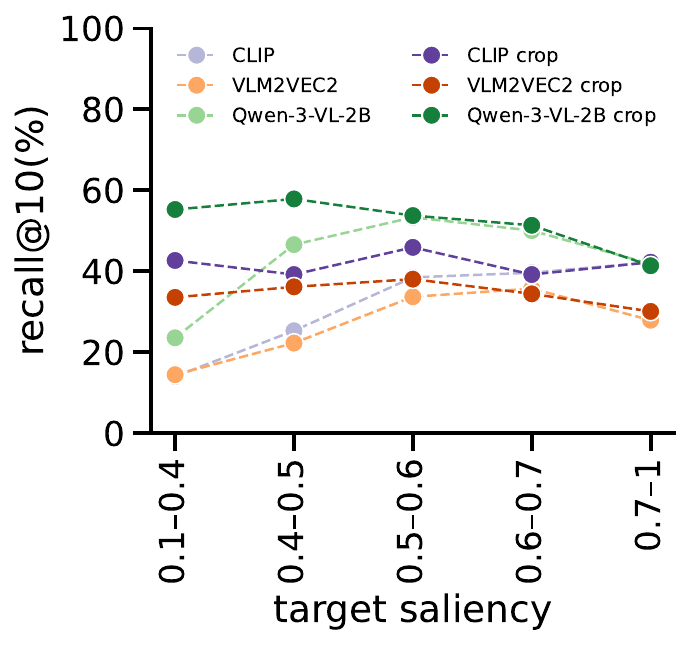}
    \caption{Recall vs saliency}
    \label{fig:saliency_specialized}
\end{subfigure}
\caption{
Stage \circled{2} object identification results for specialized models.
(b) Accuracy change under input manipulations: $\Delta$ `cropping' marks the change after cropping the target object; $\Delta$ `text' adds the question for the multimodal models.
(c) Recall@10 across visual saliency bins. 
}
\label{fig:object_identification_results_specialized}
\end{figure}

\begin{itemize}
    \item[-] $\Delta$ cropping: 
    Unlike generalized models, specialized retrievers are consistently and positively impacted by cropping, especially for ambiguous, visually complex stimuli.
    \item[-] $\Delta$ text: 
    Surprisingly, providing bimodal retrievers (\texttt{VLM2Vec2}, \texttt{Qwen-3-VL}) with the original question does not improve performance, unlike in generalized models.
    This likely reflects limited linguistic grounding ability, as current multimodal retrievers are not explicitly optimized to align textual cues with precise region-level retrieval.
\end{itemize}

\paragraph{Cropping helps specialized models when targets are non-salient.} 
Lastly, we revisit the role of visual complexity and popularity, and analyze how they affect the specialized models.
Visual complexity and popularity only weakly affect the performance of specialized models (Spearman's $\rho=0.05$ for popularity, $r_{pb}=0.13$ for visual saliency) (described in Appendix~\ref{app:image_retriever}), yet further inspection of the relation between recall and visual saliency reveals a more subtle trend, as depicted in \Cref{fig:saliency_specialized}: the more non-salient a target object is, the more cropping helps.

\begin{tcolorbox}[colback=LimeGreen!10!white, colframe=LimeGreen!75!black,left=1mm,right=1mm]
Intermediate takeaways based on object identification (stage \circled{2}):
\begin{enumerate}[topsep=0pt]
    \item All VLMs struggle with object identification, a prerequisite for successful KI-VQA. \texttt{CLIP} appears relatively strong considering its small size;
    \item Whereas VLMs do best when receiving rich multimodal input (full image and question), specialized image retrievers perform best when only receiving a cropped target object image; 
    \item Visual complexity plays less of a role than in stage \circled{1}, whereas target popularity acts as a critical bottleneck—affecting VLMs far more acutely than image retrievers.
\end{enumerate}
\end{tcolorbox}

\section{Stage \circled{3} Knowledge Retrieval and Reasoning} \label{sec:rag}

The final KI-VQA stage requires synthesizing an answer by \textbf{reasoning over knowledge}.
We assess this stage through end-task performance, while providing an analysis that isolates it from previous stages using `text-only query' metadata. 
Below, we detail the experimental setup before elaborating on our findings.

\subsection{Experimental Preliminaries}
\paragraph{Models and RAG pipeline.}
We evaluate the same suite of \textbf{generalized} VLMs as in \Cref{sec:grounding_models}. These models are first tested in a zero-shot \textit{baseline configuration}, relying solely on their parametric knowledge to answer the KI-VQA questions.

\begin{figure}[t]
    \centering
    \includegraphics[width=\linewidth]{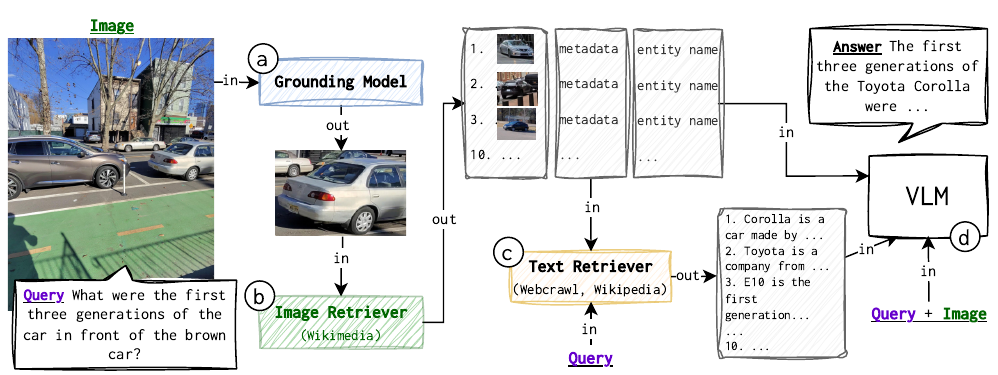}
    \caption{Grounded bimodal RAG: In stage \circled{3}, we study a modeling pipeline that combines VLMs with RAG and visual grounding. Adding the grounding module is meant to make the image retriever more effective.
    }
    \label{fig:advanced_rag}
\end{figure}

We then equip these VLMs with \textbf{RAG}, utilizing text and image retrievers as established in the original \texttt{CRAG-MM} framework \cite{cragMM2025}.
Given an input image, the image retriever returns similar images and structured metadata from a knowledge graph using \texttt{CLIP-ViT-Large-Patch14-336} \cite{Radford2021LearningTV}.
The text retriever similarly returns webpages given an input text, using documents embedded by \texttt{BGE-large-en-v1.5}~\cite{xiao2024c}. We employ the original \texttt{CRAG-MM} text and image indices.

Standard RAG can be limited by imperfect image retrieval (as observed in \Cref{sec:identification}).
Informed by \Cref{sec:grounding,sec:identification}, we evaluate a grounded bimodal RAG configuration (illustrated in \Cref{fig:advanced_rag}) that integrates an upstream visual grounding module to improve retrieval similar to region-aware retrieval\cite{lin2022revive, Jian2024LargeLM, kim2026pixel}.\footnote{Direct empirical comparison to related frameworks, such as~\cite{lin2022revive,Jian2024LargeLM,kim2026pixel}, is precluded by unavailable public codebases or incomplete implementation details; however, they represent compelling directions for future benchmarking.}
This pipeline follows these steps:
\begin{enumerate}[label=\protect\circled{\alph*}, leftmargin=*, nosep]
\item \textit{Language-based visual grounding}: \texttt{Grounding-DINO} generates a precise crop of the target region to eliminate visual noise.
\item \textit{Image retrieval}: The retriever picks images and metadata for cropped target.
\item \textit{Text retrieval}: The text retriever fetches passages using the query and the metadata from the visual retrieval step.
\item \textit{Multimodal reasoning}: The VLM generates the final answer by synthesizing the query, the original image, and the retrieved context.
\end{enumerate}

\paragraph{Evaluation metric.}
Following \cite{cragMM2025}, we evaluate QA accuracy using \texttt{gpt-4o-mini} as a binary classifier given the question, ground-truth, and prediction (prompt in Appendix~\ref{app: evaluation}).\footnote{\cite{cragMM2025} reports a 99.1\% human agreement rate for this method.} Reported scores are averaged over three runs of the judge.

\subsection{Analyses and Findings}
\begin{figure*}[t!]
\small
\centering
\begin{subfigure}[b]{0.45\textwidth}
    \centering
    \includegraphics[width=\linewidth]{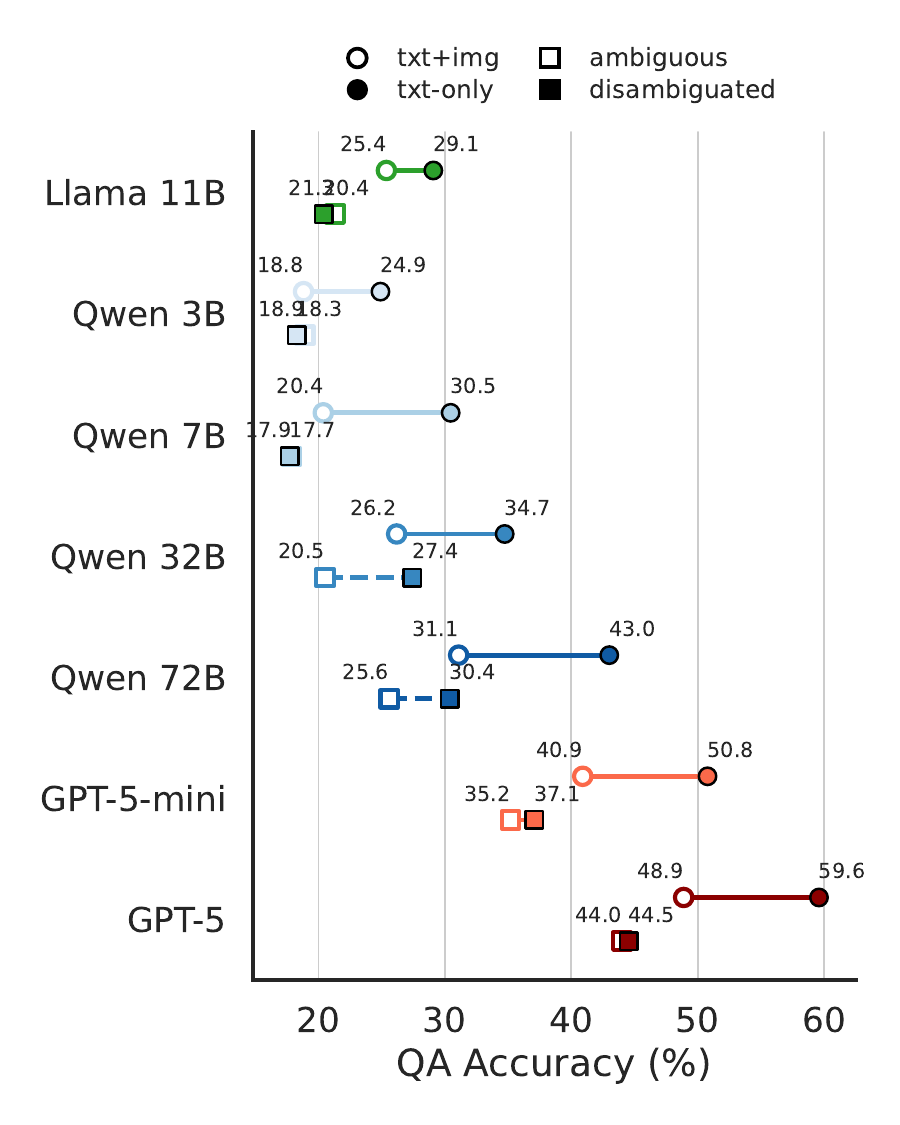}
    \caption{Generalized models}
    \label{tab:performance_baseline}
\end{subfigure}
\hfill
\begin{subfigure}[b]
{0.45\textwidth}
    \centering
    \renewcommand{\arraystretch}{0.9}
    \begin{adjustbox}{width=0.91\linewidth}
        \begin{tabular}{@{} r p{0.5em} c c c p{1em} *{2}{S} p{1em} *{2}{S} @{}}

    \toprule
\multicolumn{5}{c}{\textit{Model}} && \multicolumn{2}{c}{\textit{All}} && \multicolumn{2}{c}{\textit{Subset}} \\
\cmidrule(lr){1-5}\cmidrule(lr){7-8}\cmidrule(lr){10-11}
&& {\rotatebox[origin=c]{90}{\scriptsize ground.}}
& {\rotatebox[origin=c]{90}{\scriptsize img ret.}}
& {\rotatebox[origin=c]{90}{\scriptsize txt ret.}}
&& {\rotatebox[origin=c]{90}{\scriptsize txt+img}}
& {\rotatebox[origin=c]{90}{\scriptsize txt-only}}
&& {\rotatebox[origin=c]{90}{\scriptsize ambig.}}
& {\rotatebox[origin=c]{90}{\scriptsize dis.}} \\
\midrule

\multirow{6}{*}{\rotatebox[origin=c]{90}{\texttt{Qwen 32B}}}
&& \xmark & \xmark & \xmark && 26.2 & 34.7 && 20.5 & 27.4 \\
\cmidrule{3-11}
&& \xmark & \cmark & \xmark && 27.2 & {{}-} && 22.7 & 27.5 \\
&& \xmark & \xmark & \cmark && 31.0 & 57.2 && 25.9 & \bfseries 33.6 \\
&& \xmark & \cmark & \cmark && 34.4 & {{}-} &&  24.5 &  29.6 \\ \cmidrule{3-11}
&& \cmark & \cmark & \cmark && \textbf{34.7} & {{}-} && \textbf{28.0} & 29.9 \\ 
\rowcolor{lightgray}  && \texttt{GT} & \cmark & \cmark && 36.1 & {{}-} && 32.8 & 33.6\\
\midrule

\multirow{6}{*}{\rotatebox[origin=c]{90}{\texttt{GPT-5}}}
&& \xmark & \xmark & \xmark && 48.9 & 59.6 && 44.0 & 44.5 \\
  \cmidrule{3-11}
&& \xmark & \cmark & \xmark && 61.5 & {{}-} && 50.9 & 53.9 \\
&& \xmark & \xmark & \cmark && 56.7 & 73.3 && 53.1 & \bfseries 57.1 \\
&& \xmark & \cmark & \cmark && 61.5 & {{}-} && 46.4 & 55.2 \\  \cmidrule{3-11}
&& \cmark & \cmark & \cmark && \bfseries 62.2 & {{}-} && \bfseries 53.9 & 56.5 \\
\rowcolor{lightgray} && \texttt{GT} & \cmark & \cmark && 62.2 & {{}-} && 55.7 & 58.4 \\
\bottomrule
    
    \end{tabular}
    \end{adjustbox}
    \caption{Ablations on grounded RAG components}
    \label{tab:performance_rag}
\end{subfigure}
\caption{
Comparison of (a) generalized models and (b) RAG pipeline ablations on KI-VQA accuracy. \textit{All} reports performance over all instances; \textit{Subset} evaluates ambiguous (\textit{ambig.}) vs. disambiguated (\textit{dis.}) queries. Columns \textit{ground.}, \textit{img ret.}, and \textit{txt ret.} denote the activation of grounding, image retrieval, and text retrieval modules. \textit{txt-only} evaluates questions with gold entity names replacing image, while \textit{txt+img} uses the standard VQA setup. Shaded rows indicate use of ground-truth (GT) target regions.
}
\label{tab:end-task-accuracy}
\end{figure*}

The end-task accuracies in \Cref{tab:end-task-accuracy} underscore the difficulty of out-of-the-box KI-VQA, even for proprietary models (see `txt+img' in \Cref{tab:performance_baseline}). Although ambiguous expressions (`ambiguous' marker) degrade performance, the modest 1.8-point gain from disambiguation suggests ambiguity is not the primary bottleneck.

\paragraph{Retrieval/reasoning is the primary KI-VQA bottleneck.}
Using the `text-only questions' from \dataset---where target entity names are explicitly mentioned---decouples retrieval and reasoning from stage \circled{1} and \circled{2} localization errors. \Cref{tab:performance_baseline} (`txt-only' marker) shows that while this improves results by 8.7 points on average, the majority of errors persist. Notably, \texttt{GPT-5} improves most, with 20.9\% of errors disappearing. This underscores that parametric knowledge is insufficient for most KI-VQA tasks, particularly for smaller models with limited capacity.

\paragraph{Unimodal RAG yields modest gains.} 
\Cref{tab:performance_rag} demonstrates the impact of coupling VLMs with image and text retrievers. Image retrieval provides a consistent 6.8-point average boost for \texttt{Qwen} and \texttt{GPT-5}. Additionally, text retrieval alone adds 6.3 points; however, its effectiveness is limited by uninformative queries (e.g., ``Is this sold in South America?'' in \Cref{fig:pipeline_illustration}). By using `text-only' questions with explicit entity names (replacing ``white vehicle'' with ``Tesla Model Y'' in ``How does the range of the white vehicle compare to that of the hyundai ioniq 5?''), performance jumps by 16.6 and 26.2 points for \texttt{GPT-5} and \texttt{Qwen}, respectively. These results emphasize that accurate query reformulation and object identification are critical for maximizing retrieval gains.

\paragraph{Bimodal and grounded RAG yield top scores.}
Combining bimodal retrieval with visual grounding achieves our highest performance (\Cref{tab:performance_rag}), confirming that `cleaning' visual inputs affects the KI-VQA positively. Using ground-truth regions (\texttt{GT}, highlighted in gray) provides an additional 1.4-point gain for \texttt{Qwen}, yet performance still trails the text-only retrieval baseline. Since image retrieval is implicitly intended to serve object identification, this reinforces that it does not succeed in that task, even when given a ground-truth target region.
The error analysis in Appendix~\ref{app:error_analysis} shows that 25\% of failures stem from poor object identification even with \texttt{GPT-5}. This highlights the need for more robust recognizers handling fine-grained taxonomies and effective retrieval to bridge parametric knowledge gaps.

\begin{tcolorbox}[colback=Apricot!10!white, colframe=Apricot!75!black,left=1mm,right=1mm]
Intermediate takeaways from the knowledge retrieval and reasoning stage:
\begin{enumerate}[topsep=0pt]
    \item Even proprietary models like \texttt{GPT-5} see substantial gains when target entity names are explicitly provided, indicating that internal knowledge cannot substitute for effective retrieval in KI-VQA;
    \item Integrating a visual grounding module to crop target objects before retrieval consistently yields the highest performance, demonstrating that `cleaning' the visual input is vital for accurate multimodal RAG;
    \item Even with ground-truth target regions, performance lags behind text-only retrieval baselines, revealing that current retrieval pipelines still fail at fine-grained object identification and reasoning.
\end{enumerate}
\end{tcolorbox}
\section{Conclusion}
\label{sec:conclusion}
We introduced \dataset, a diagnostic benchmark designed for the stage-wise evaluation of KI-VQA under realistic visual conditions.
By augmenting \texttt{CRAG-MM} with fine-grained annotations---including bounding boxes, referring expression types, disambiguated or text-only queries, and visual saliency metadata---we conducted a systematic analysis of the KI-VQA pipeline. Our analysis spanned fully parametric VLMs, specialized grounding and retrieval models, and modular retrieval-augmented pipelines.

Our findings surface many patterns that explain why KI-VQA is complex, such as the visual complexity of the scene, the popularity of the target object, and the complexity of the referring expression directly correlating with performance in the various stages of the pipeline.
We also uncovered failure modes (such as the surprising find that \texttt{GPT-5} struggles much more with explicit grounding than \texttt{Qwen}) and determined that for some stages, a small, specialized model can outperform larger, generalized VLMs.
We end with the following three concrete lessons for the design of future KI-VQA systems:

\begin{enumerate}
    \item \textbf{Linguistic ambiguity is a structural problem, not a minor nuisance.} Disambiguating referring expressions yields consistent gains across all stages of the pipeline. Future benchmarks should audit and annotate query ambiguity, and models should explicitly address it as ignoring ambiguity will lead to consistent underestimation of models' reasoning capacity. We advocate for models that explicitly detect and resolve ambiguity—potentially through iterative query reformulation—prior to knowledge retrieval.

    \item \textbf{Object identification should inform text retrieval.} Text retrieval with original, image-dependent queries is often uninformative when the textual question alone does not identify the target (e.g., ``Is this car sold in South America?''). Using the target entity name in the retrieval query---derived from either model prediction or early-stage identification---yields the largest performance gains observed in our study. This `identify first, then retrieve' ordering is the most impactful design principle we uncover.

    \item \textbf{Knowledge retrieval and reasoning is the dominant bottleneck.} Critically, even when models are provided the ground-truth entity name, the majority of errors persist, indicating that limited parametric knowledge and imperfect reasoning---not visual grounding or object identification---account for most KI-VQA failures. This suggests that the field should invest more in improving retrieval quality and multi-hop reasoning over retrieved evidence.
\end{enumerate}

By providing a foundation for \emph{stage-aware} KI-VQA evaluation, \dataset encourages a shift from `black-box' testing to principled, modular assessment. We hope this benchmark serves as a catalyst for developing multimodal knowledge assistants that can truly navigate the intersection of visual perception and world knowledge.


\section*{Acknowledgements}
We thank the reviewers for their valuable suggestions. We thank our colleagues from McGillNLP and Mila, especially Marius Mosbach, Vaibhav Adlakha, Rabiul Awal, and Aishwarya Agrawal.
PB is supported by the RBC Borealis AI Global Fellowship Award and the ServiceNow-Mitacs Accelerate program.
SR is supported by the Canada CIFAR AI Chair, the NSERC Discovery Grant and the Sloan Fellowship. The project is partly funded by the IVADO R3AI program.
VD acknowledges the support of the IVADO Postdoctoral Research Funding.

%
%
\clearpage
\bibliographystyle{splncs04}
\bibliography{main}

\clearpage
\appendix
\section{Dataset}\label{app:dataset}
\subsection{Dataset Comparison}

\begin{table}[t]
\centering
\caption{Comparison of relevant VQA benchmarks. \textbf{Knowledge} denotes the type of knowledge required (`-' denotes no explicit requirement); \textbf{Non-salient} marks whether a benchmark focuses on non-salient targets; \textbf{Decoupled} indicates whether stage-wise evaluation is performed (\textbf{G}rounding, \textbf{O}bject Identification, \textbf{Q}uestion Answering); \textbf{Ambiguity} indicates whether or not ambiguity effects are evaluated.\protect\footnotemark}
\label{tab:dataset_comparison_final}
\resizebox{\columnwidth}{!}{%
\begin{tabular}{@{} l c c c c @{}}
\toprule
\textbf{Dataset} & \textbf{Knowledge} & \textbf{Non-salient} & \textbf{Decoupled Analysis} & \textbf{Ambiguity} \\ \midrule
\textit{Knowledge-Intensive VQA} & & & & \\
OK-VQA \cite{marino2019ok} & Commonsense & \xmark & \xmark (Q only)& \xmark \\
A-OKVQA \cite{Schwenk2022AOKVQAAB} & Reasoning & \xmark & \xmark (Q only) & \xmark \\
EncyclopedicVQA \cite{Mensink2023EncyclopedicVV} & Factual & \xmark & \xmark (Q only) & \xmark \\
SnapNTell \cite{qiu2024snapntell} & Factual & \xmark & \xmark (Q only) & \xmark \\
ViQuAE \cite{Lerner2022ViQuAEAD} & Factual & \xmark & \xmark (Q only) & \xmark \\
ReasonVQA \cite{tran2025reasonvqa} & Reasoning & \xmark & \xmark (Q only) & \xmark \\
DynVQA \cite{li2025benchmarking} & Dynamic & \xmark & \xmark (Q only) & \xmark \\ 
InfoSeek \cite{chen2023can} & Factual & \xmark & \xmark (Q only) & \xmark \\
OVEN \cite{Hu2023OpendomainVE} & - & \xmark & \xmark (O only) & \xmark \\
\midrule
\textit{Realworld Perception} & & & & \\
V* Benchmark \cite{wu2024v} & - & \cmark & \xmark (Q only) & \xmark \\
GigaGrounding \cite{Ma2024WhenVG} & - & \cmark & \xmark (G only) & \xmark \\
MME-RealWorld \cite{zhangmme} & - & \cmark & \xmark (Q only) & \xmark \\
MRAG-Bench \cite{hu2025mragbench} & Visual & \cmark & \xmark (Q only) & \xmark \\
CRAG-MM \cite{cragMM2025} & Factual & \cmark & \xmark (Q only) & \xmark \\ \midrule
\textit{Diagnostic/Capability} & & & & \\
VisualSimpleQA \cite{Wang2025VisualSimpleQAAB} & Factual & \cmark & Visual and Linguistic module (Q only) & \xmark \\
HallusionBench \cite{guan2024hallusionbench} & - & \xmark & Visual, Language Hallucination & \xmark \\
Prism \cite{qiao2024prism} & - & \xmark & Perception and reasoning & \xmark \\
MMVet \cite{yu2024mm} & - & \xmark & Capability Integration  & \xmark \\ \midrule
\textbf{\dataset (Ours)} & Factual & \cmark & Stage-wise evaluation for \textbf{G, O, Q} & \cmark \\ \bottomrule
\end{tabular}%
}
\end{table}

\Cref{tab:dataset_comparison_final} compares relevant VQA benchmarks with \dataset.

\footnotetext{
Although prior work has studied ambiguity in general VQA settings \cite{bhattacharya2019does,stengel2023did,ni2025visualo}, we focus on a underexplored ambiguity in KI-VQA—errors in precise entity identification and visual grounding that occur before knowledge retrieval.
}

\begin{figure}[t]\centering
    \includegraphics[width=0.33\linewidth]{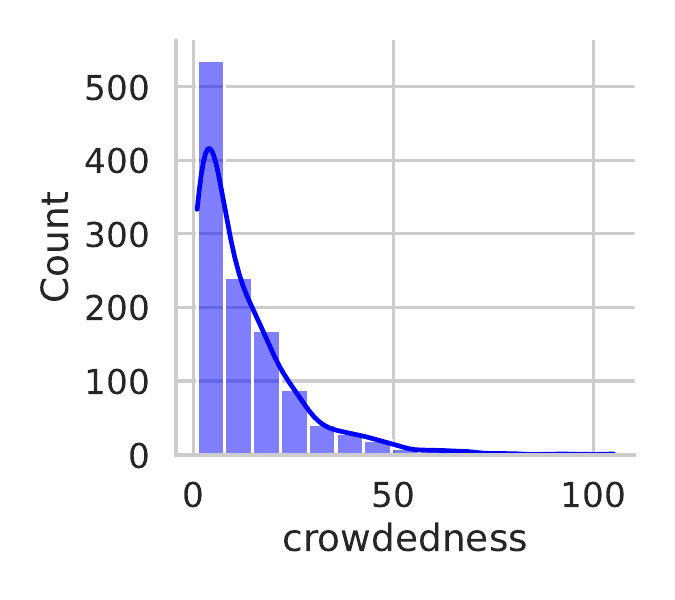}
    \includegraphics[width=0.33\linewidth]{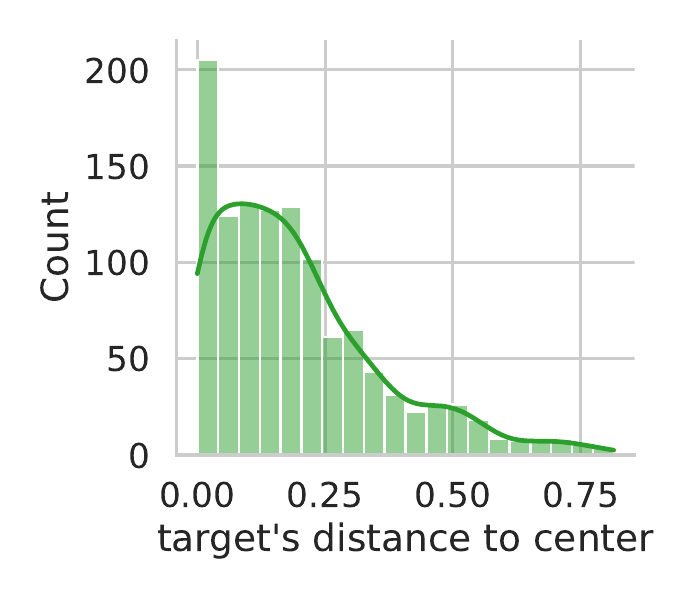}
    \includegraphics[width=0.33\linewidth]{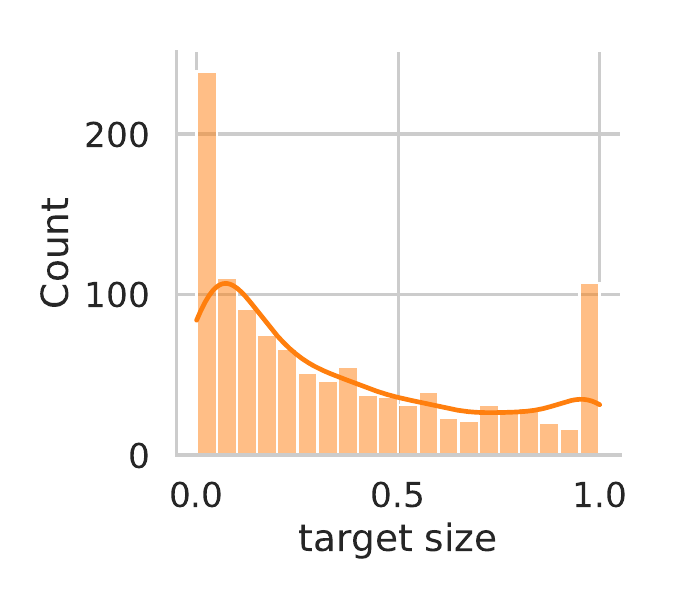}
    \includegraphics[width=0.33\linewidth]{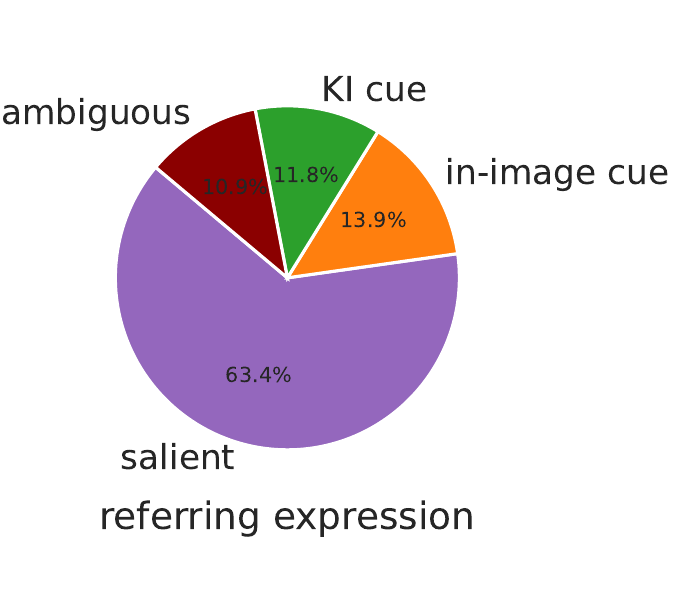}
    \caption{\dataset statistics based on scene crowdedness, the target object's distance to center, the target's size, and the referring expression types.}
    \label{fig:dataset-stats}
    \vspace{-0.3cm}
\end{figure}
\subsection{Dataset Statistics} \label{app:dataset-stats}
\Cref{fig:dataset-stats} presents the statistics of the \dataset samples with additional annotations.
Note that, from the \texttt{CRAG-MM} dataset, we remove examples that (a) are not actually \textit{knowledge-intensive}, based on \texttt{CRAG-MM}'s `simple-recognition' tag, (b) concern dynamically changing knowledge, based on the `fast-changing' and `real-time' tags, and (c) concern domains which do not allow us to tag a specific entity visually (this concerns the domains `text understanding', `math \& science', `book', `shopping' and `food').

In the main paper, we broke down performance for different intervals for visual complexity metrics and the popularity metric. We did that by performing interval grouping and using similar subgroup sizes for each metric across the dataset, with each subgroup comprising approximately 230 instances ($\approx$20\% of the total pool).

\subsection{Dataset Examples} \label{app:dataset_examples}

\Cref{tab:example} presents representative examples from the dataset across different types of referring expressions, along with the metadata collected during the human annotation stage for \dataset.

\subsection{Metadata Annotation} \label{app:dataset_metadata}
This section includes details about the additional metadata added to \dataset.

\paragraph{Preliminary annotations} To evaluate intermediate stages such as visual grounding and object identification, we build an automatic pipeline to collect target entity names, Wikipedia URLs, and bounding box annotations aligned with the original (image, question, answer) triplets. 
Target entities are extracted from the answer using the spaCy entity linker~\cite{honnibal2020spacy}. Bounding box annotations are obtained via a two-stage approach: (1) we extract a referring expression for the target object using Llama-3.2-11b-Vision-Instruct~\cite{meta2024llama}, given the image and query; (2) we apply Grounding-DINO-Base ~\cite{Liu2023GroundingDM} with a text and box threshold of 0.1 to localize the corresponding image regions. Yet, note that this annotation is only preliminary and that the human annotator should verify and potentially modify entity names, Wikipedia URLs, and bounding boxes.

\noindent\begin{minipage}{\textwidth}
\centering
\captionof{table}{Examples from \dataset for different types of referring expressions, along with various types of metadata we collect.}
\setlength{\tabcolsep}{8pt}
\resizebox{1.0\linewidth}{!}{
\begin{tabular}{m{0.1cm}m{3.5cm}|m{6.5cm}m{6.5cm}}
        \toprule
         \multicolumn{2}{c|}{Referring Expression} & \multicolumn{1}{c}{\textbf{Salient}}&\multicolumn{1}{c}{\textbf{Ambiguous}} \\ 
        \midrule
        \multicolumn{2}{c|}{Image} &  \centering \includegraphics[width=45mm,height=45mm]{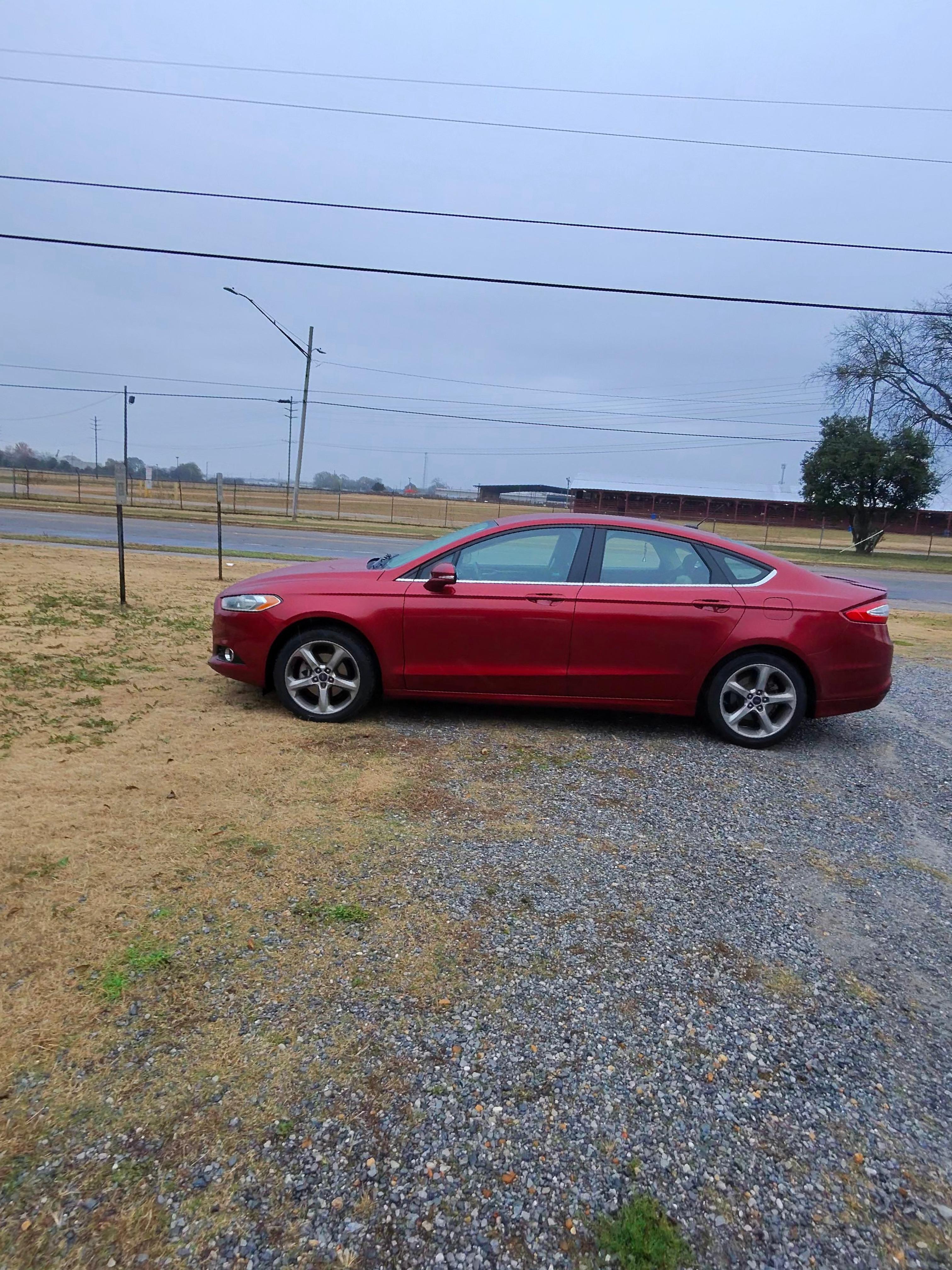} & \centering \includegraphics[width=45mm,height=45mm]{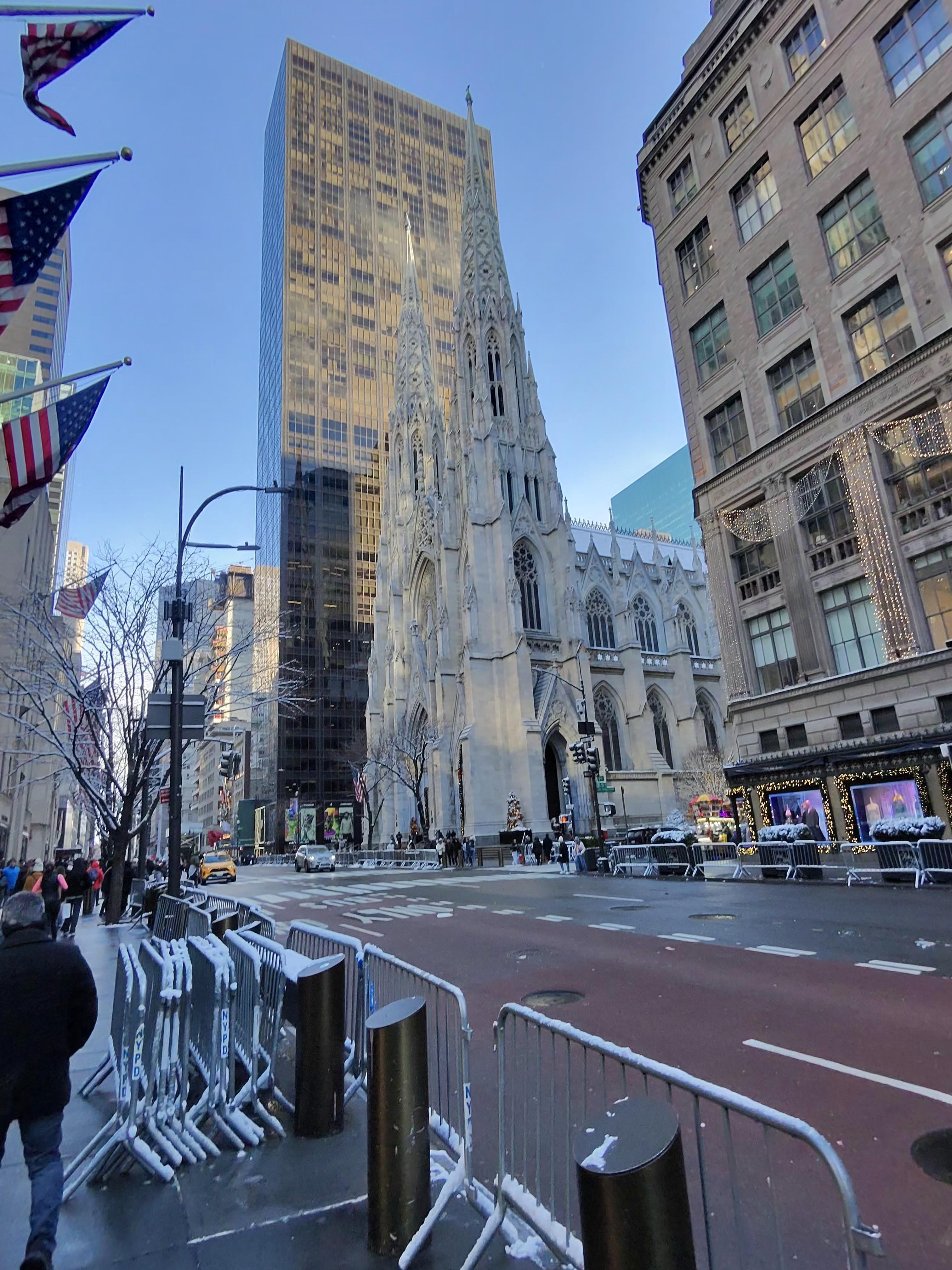}  \tabularnewline
        \midrule 
        \multicolumn{2}{c|}{Question} & which came out first, \underline{\textbf{this model}} or the ford focus? & was the empire state building around when \underline{\textbf{this}} was first built?\\
        \midrule
        \multicolumn{2}{c|}{Answer}& the ford focus, introduced in 1998, came out before the ford fusion, introduced in 2006. & no, the empire state building was not around when st.\ patrick's cathedral was first built, as the cathedral was completed in 1878 and the empire state building was built from 1930 to 1931.\\
        \midrule
        \multirow{7}{*}[-3em]{\rotatebox[origin=c]{90}{\textbf{Metadata}}}
         & Size of Target Obj. & 0.1411 & 0.2419 \\ 
         \cmidrule{2-4}
         & Target ROI Box & [643,1610,2854,2388]  & [1136, 256, 2320, 2747]\\ 
         \cmidrule{2-4}
         & Target Obj. Name  & Ford Fusion (Americas) & St.\ Patrick's Cathedral (New York City)\\ 
         \cmidrule{2-4}
         & Target Obj. Wiki URL & 
         \href{https://en.wikipedia.org/wiki/Ford_Fusion_(Americas)}{en.wikipedia.org/Ford Fusion (Americas)}  & \href{https://en.wikipedia.org/wiki/St._Patrick's_Cathedral_(New_York_City)}{en.wikipedia.org/St.Patrick's Cathedral}\\ 
         \cmidrule{2-4}
         & Textual Only Question & which came out first, the ford fusion or the ford focus?  & was the empire state building around when st.\ patrick's cathedral was first built? \\ 
         \cmidrule{2-4}
         & Popularity & 296969& 203644\\ 
         \cmidrule{2-4}
         & Disambig. Query & - & was the empire state building around when \underline{\textbf{this cathedral}} was first built?\\
         \bottomrule
         \addlinespace[10pt]
         \toprule
         \multicolumn{2}{c|}{Referring Expression} &\multicolumn{1}{c}{\textbf{Knowledge-Intensive Cue}} &\multicolumn{1}{c}{\textbf{In-Image Cue}} \\ 
        \midrule
        \multicolumn{2}{c|}{Image} &   \centering \includegraphics[width=45mm,height=45mm]{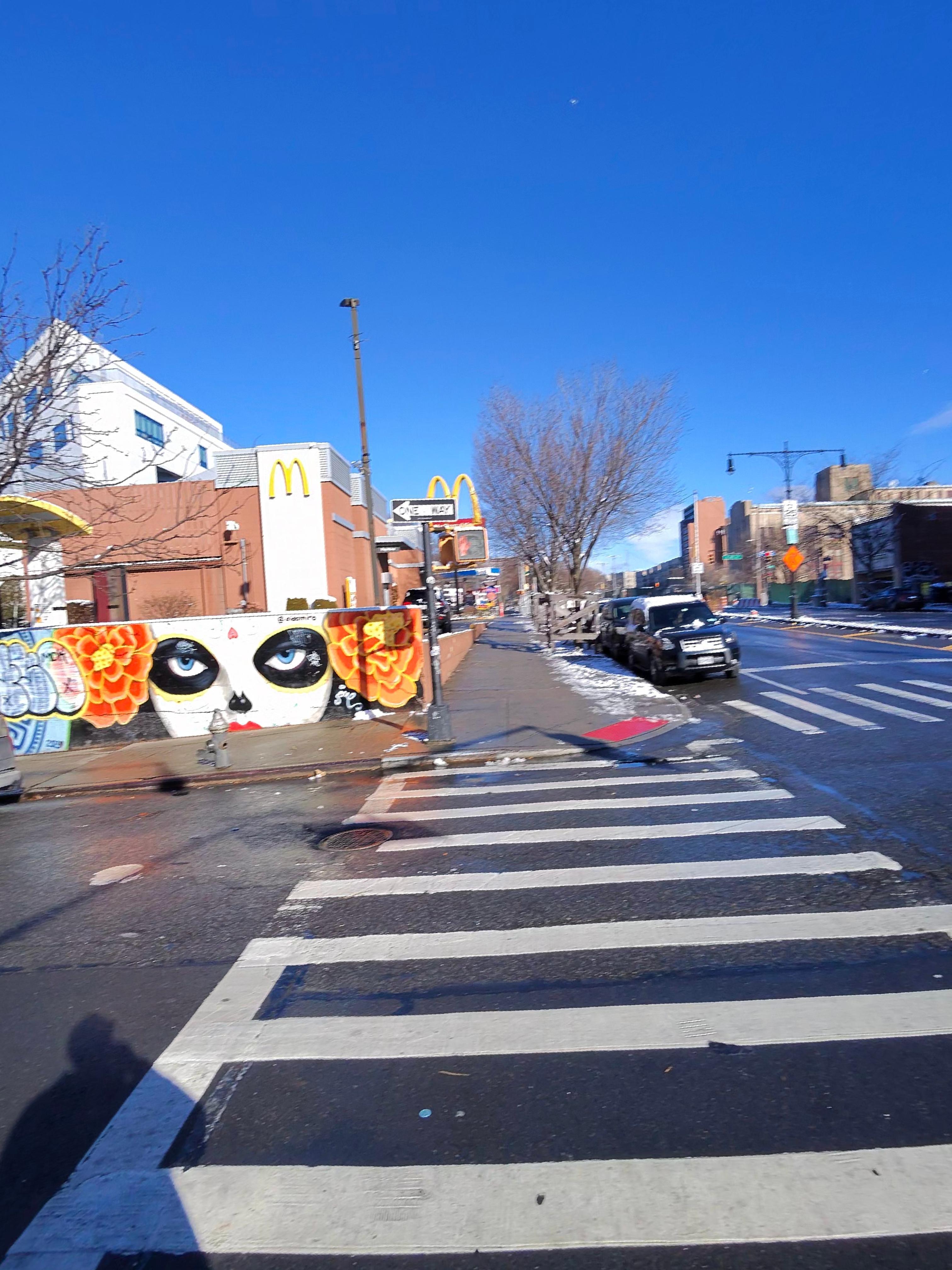} &  \centering \includegraphics[width=45mm,height=45mm]{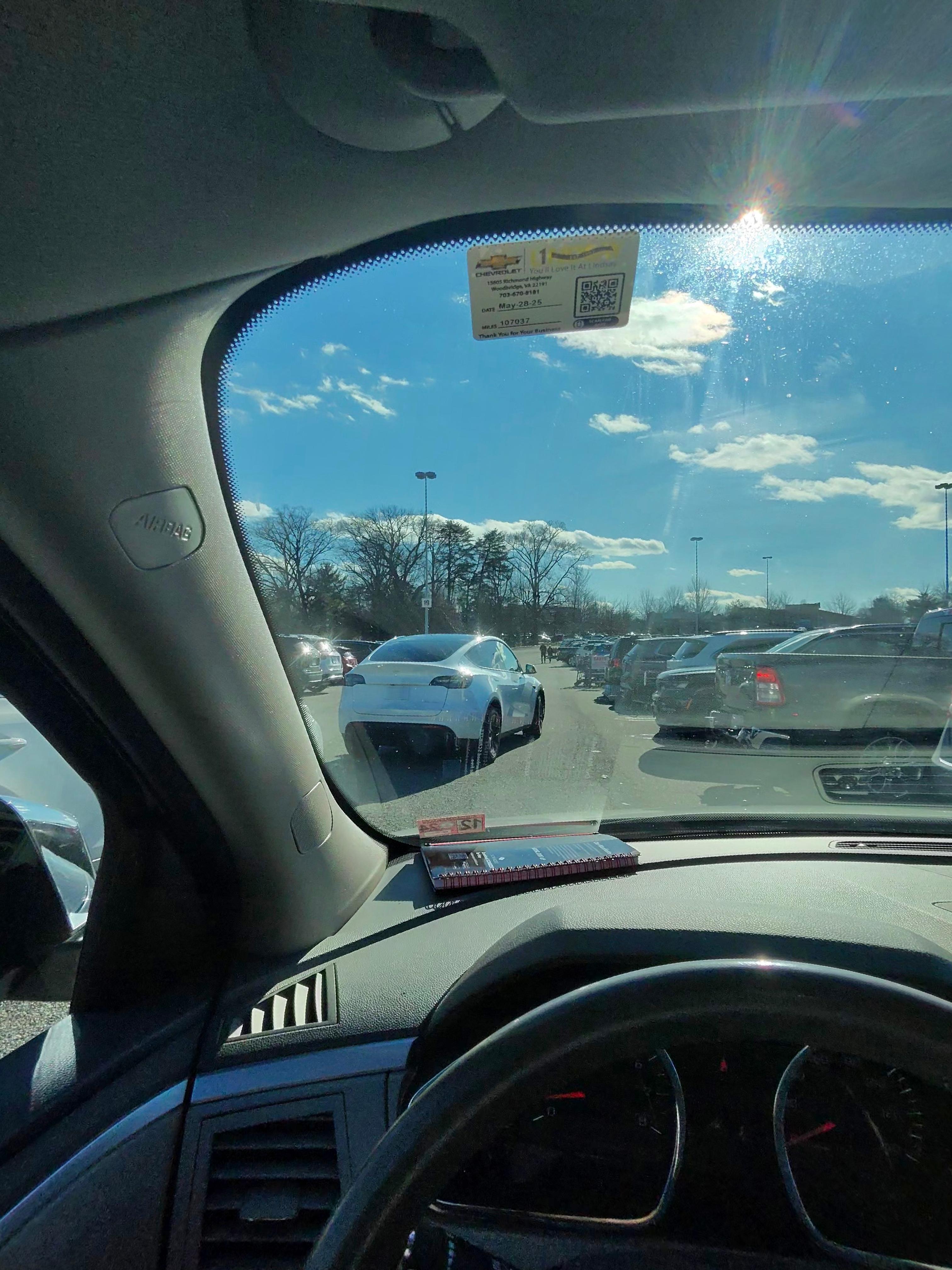}  \tabularnewline
        \midrule 
        \multicolumn{2}{c|}{Question} & how is \underline{\textbf{this restaurant's logo}} color different from wendy's logo? & how does the range of the \underline{\textbf{white vehicle}} compare to that of the hyundai ioniq 5?\\
        \midrule
        \multicolumn{2}{c|}{Answer}&  mcdonald's logo is primarily yellow and red, while wendy's logo is primarily red and white. & the tesla model y typically offers a slightly longer range than the hyundai ioniq 5, with the model y long range reaching up to 330 miles and the ioniq 5 limited up to 303 miles, though the ioniq 5 charges faster. \\
        \midrule
        \multirow{7}{*}[-3em]{\rotatebox[origin=c]{90}{\textbf{Metadata}}}
         & Size of Target Obj. &  0.0649 & 0.0245\\ 
         \cmidrule{2-4}
         & Target ROI Box  & [2,1400,1345,1989] & [1071, 2003, 1736, 2452]\\ 
         \cmidrule{2-4}
         & Target Obj, Name  & McDonald's & Tesla Model Y \\ 
         \cmidrule{2-4}
         & Target Obj. Wiki URL  & \href{https://en.wikipedia.org/wiki/McDonald's}{en.wikipedia.org/McDonald's} &\href{https://en.wikipedia.org/wiki/Tesla_Model_Y}{en.wikipedia.org/Tesla Model Y} \\ 
         \cmidrule{2-4}
         & Textual Only Question  & how is mcdonald's logo color different from wendy's logo? & how does the range of the tesla model y compare to that of the hyundai ioniq 5? \\ 
         \cmidrule{2-4}
         \cmidrule{2-4}
         & Popularity &2524618  &650057 \\ 
         \cmidrule{2-4}
         & Disambig. Query  & - & -\\
        \bottomrule
\end{tabular}
}
\label{tab:example}
\end{minipage}

\begin{figure}[t]
    \centering
    \begin{minipage}{0.5\linewidth}
        \centering
        \includegraphics[width=\linewidth]{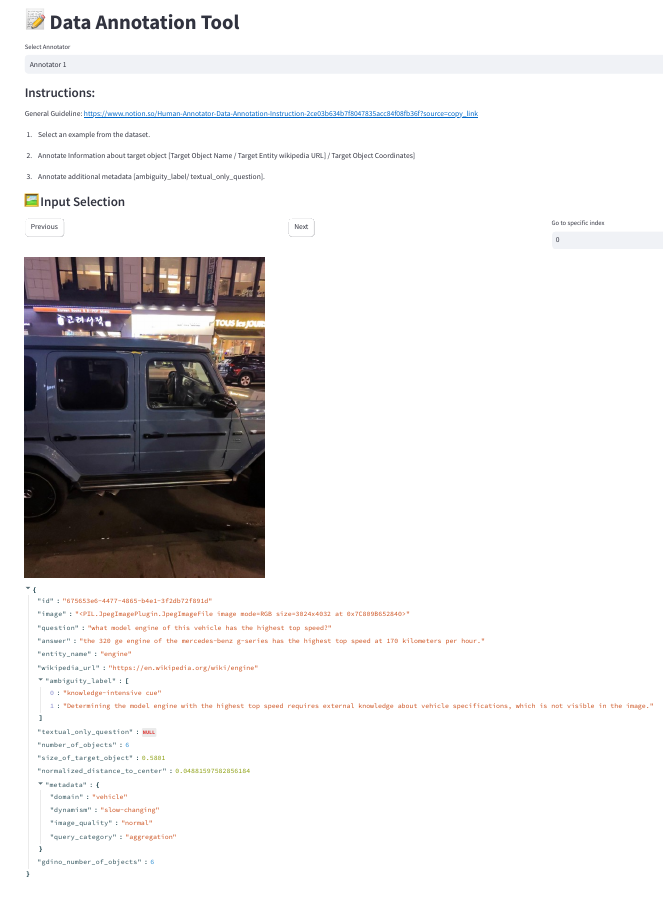}
    \end{minipage}
    \hfill
    \begin{minipage}{0.45\linewidth}
        \centering
        \includegraphics[width=\linewidth]{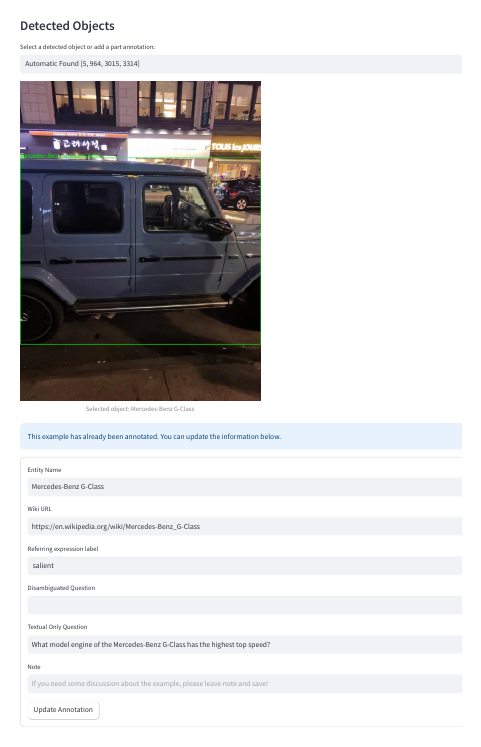}
    \end{minipage}
    \caption{Human annotation interface.}
    \label{fig:annotation_interface}
\end{figure}

\paragraph{Human Annotation Procedure and Tool}\label{app:dataset:annotation}
To refine and verify the preliminary annotations as well as to generate additional metadata, we employ an annotation tool, as illustrated in Figure~\ref{fig:annotation_interface}. The main verification process involves three steps: 1) annotating the target region of interest (ROI) for each multimodal input 
(i.e., query and image), 2) mapping the target object to its corresponding Wikipedia entity, including the associated Wikipedia URL; and 3) checking whether the question is ambiguous or lacks sufficient clues to uniquely identify the target within the example.

The annotation procedure is conducted in two stages. In the first stage, five annotators independently annotate the dataset using the provided annotation tool (Figure~\ref{fig:annotation_interface}). In the second stage, two annotators review the annotations to identify potential errors and ensure quality. Importantly, reviewers are assigned instances different from those they annotated in the first stage, ensuring that no annotators review their own annotations.

\paragraph{Distance from the center of the image of the target object.}
The remaining types of metadata for visual complexity can be computed based on the ROI bounding boxes.
We compute the distance between the image center and the center of a ROI, normalized by the image size.

Let the image have width $W$ and height $H$. The image center is
\begin{equation}
C_{\text{img}} = \left( \tfrac{W}{2}, \tfrac{H}{2} \right).
\end{equation}

An ROI is defined by its bounding box $(x_{\min}, y_{\min}, x_{\max}, y_{\max})$, 
and its center is given by
\begin{equation}
C_{\text{roi}} = \left( \tfrac{x_{\min} + x_{\max}}{2}, \; 
                     \tfrac{y_{\min} + y_{\max}}{2} \right).
\end{equation}

The Euclidean distance between the centers is
\begin{equation}
d = \sqrt{\Big(\tfrac{W}{2} - \tfrac{x_{\min} + x_{\max}}{2}\Big)^{2} \;+\;
          \Big(\tfrac{H}{2} - \tfrac{y_{\min} + y_{\max}}{2}\Big)^{2}}.
\end{equation}

We normalize this distance by half of the image diagonal length. 
The image diagonal is
\begin{equation}
D = \sqrt{W^{2} + H^{2}},
\end{equation}
so the half-diagonal is $D/2$. The normalized distance is therefore
\begin{equation}
d_{\text{norm}} = \frac{d}{D/2} \;=\; \frac{2d}{D}.
\end{equation}
This normalized score lies in $[0,1]$, where $d_{\text{norm}} = 0$ indicates 
perfect alignment of the ROI center with the image center, and 
$d_{\text{norm}} = 1$ corresponds to maximum displacement along the half-diagonal.

\paragraph{Number of Objects in an Image.} To quantify the visual clutter of an image, which increases the visual complexity of model recognition for a given query, we utilize the number of objects detected by the object detection model Grounding DINO \cite{Liu2023GroundingDM}. We empirically find that a standard closed-vocabulary object detector with 81 predefined categories, such as YOLO~\cite{redmon2016you}, cannot cover all target objects in our dataset; thus, we use an open-vocabulary detection model here. With 81 predefined object categories set by YOLO, we additionally add more categories to cover objects in our dataset, such as `building', `tower', `plant' and `gardening'.

\paragraph{Proportion of Target Size.} Given the bounding box of the target entity \hfill \\ 
$(x_{\min}, y_{\min}, x_{\max}, y_{\max})$, we compute its area as $A_{\text{roi}} = (x_{\max} - x_{\min}) \times (y_{\max} - y_{\min})$. The image area is $A_{\text{img}} = W \times H$, where $W$ and $H$ denote the image width and height, respectively. The proportion of target size is then defined as the ratio $s = \frac{A_{\text{roi}}}{A_{\text{img}}}$. This normalized measure lies in $[0,1]$, where values close to $0$ indicate that 
the target occupies a very small region in the image, and values closer to $1$ 
indicate that the target covers most of the image.

\paragraph{Visual Saliency}
We define a saliency score inspired by \cite{Wang2025VisualSimpleQAAB} that quantifies how visually prominent a target object is within an image by jointly considering its relative size, spatial centrality (distance to the center), and scene clutter (crowdedness). Intuitively, an object is more salient if it is larger, closer to the image center, and appears in less cluttered scenes.
Let $s$, $d$, and $n$ denote the target object’s size, its distance from the image center, and the number of detected objects in the image, respectively. Each component is first normalized to the range $[0,1]$ using a normalization function $\mathcal{N}(\cdot)$ (min–max or robust normalization):
\begin{equation}
\mathcal{N}(x) = \frac{x - \min(x)}{\max(x) - \min(x)}.
\end{equation}
Size is treated as positively correlated with saliency, while distance and clutter are negatively correlated. Accordingly, we compute:
\begin{equation}
\tilde{s} = \mathcal{N}(s), \quad \tilde{d} = 1 - \mathcal{N}(d), \quad \tilde{c} = 1 - \mathcal{N}(\log(1+n)).
\end{equation}
The logarithmic transformation on $n$ reflects the sub-linear perceptual effect of clutter (e.g., differences between 10 and 100 objects are less salient than raw counts suggest).
The final saliency score $S \in [0,1]$ is obtained via a weighted geometric mean, which encourages balanced contributions across factors and penalizes cases where any single component is weak:
\begin{equation}
S = \exp\left( \frac{ \alpha \log(1+\tilde{s}) + \beta \log(1+\tilde{d}) + \gamma \log(1+\tilde{c}) }{\alpha + \beta + \gamma} \right) - 1,
\end{equation}
where $\alpha$, $\beta$, and $\gamma$ control the relative importance of size, centrality, and clutter. In our experiments, all weights are set to 1.0.
This formulation yields a bounded, interpretable saliency measure that is robust to scale differences across images and naturally captures the interaction between visual prominence cues.

\subsection{Subjectivity Analysis} \label{app: annotation_subjectivity}
To evaluate the potential impact of annotation subjectivity and ambiguity within our data creation pipeline, we conduct an independent human validation study. We recruit 12 professional annotators via Prolific\footnote{https://www.prolific.com/} to evaluate a stratified sample of 100 image--question pairs ($n=100$, balanced with 25 instances per referring expression category). Annotators are tasked with identifying the appropriate referring expression categories for each pair, with majority-voting applied across three independent responses per instance. 

Our analysis demonstrates robust human consensus: we achieve moderate agreement for the multi-class classification task (Cohen's $\kappa=0.453$) and substantial agreement for the binary ambiguity classification task ($\kappa=0.605$) between the expert ground-truth annotations and the crowd-sourced majority votes. These results confirm a high degree of objective human consensus, indicating that the dataset's definitions remain robust against individual annotation or author bias.
\section{Generalizability of Findings} \label{app: generalizability}

Many existing KI-VQA benchmarks present significantly simplified visual environments. For instance, datasets like InfoSeek~\cite{chen2023can} and EncyclopedicVQA (E-VQA)~\cite{Mensink2023EncyclopedicVV} exhibit low object densities, typically ranging from 1.2 to 1.6 objects per image. In contrast, CRAG-MM~\cite{cragMM2025} averages 12.5 objects per image, thereby serving as a more rigorous testbed for multimodal knowledge assistants in the wild settings. Furthermore, our insights regarding the relationship between visual complexity and model performance generalize well beyond egocentric domains, aligning closely with findings from non-egocentric benchmarks such as MRAG~\cite{hu2025mragbench} and VisualSimpleQA~\cite{Wang2025VisualSimpleQAAB}.

To further investigate whether the visual grounding module integrated into the RAG pipeline in \Cref{sec:rag} consistently boosts performance across different KI-VQA environments, we integrate it into the EchoSight \cite{yan2024echosight} framework, leveraging its prebuilt retrieval index for EVQA-based RAG. As detailed in \Cref{tab:grounding_subgroups}, while introducing the grounding module leads to an overall performance degradation on EVQA, a granular category-level analysis reveals nuanced behavior. Specifically, grounding degrades performance in ``landmark'' categories where global visual context would be inherently preferred. Conversely, performance improvements are concentrated in highly object-centric subsets (e.g., iNaturalist). These findings suggest that universal grounding is not a one-size-fits-all solution; instead, we recommend that future work explore routing architectures capable of dynamically invoking grounding modules based on query and image characteristics.

\begin{table}[h]
\centering
\small
\caption{The impact of grounding on another KI-VQA benchmark (EVQA).}
\label{tab:grounding_subgroups}
    \begin{tabular}{@{}llcc@{}}
    \toprule
    \textbf{Dataset / Subset} & \textbf{Model} & \textbf{Recall@1/5/10 (\%)} & \textbf{QA (\%)} \\ \midrule
    \multirow{2}{*}{\textbf{EVQA (All)}} & EchoSight & 13.4 / 31.8 / 41.7 & 41.7 \\
    & + Grounding & 11.8 / 29.3 / 37.6 & 39.5 \\ \cmidrule(lr){2-4}
    \multirow{2}{*}{\textit{- iNaturalist (Obj)}} & EchoSight & 9.0 / 25.4 / 34.6 & 39.0\\
    & + Grounding & 9.6 / 27.5 / 35.8 & 40.3 \\ \cmidrule(lr){2-4}
    \multirow{2}{*}{\textit{- Landmarks}} & EchoSight & 18.2 / 38.9 / 49.7 & 44.8  \\ 
    & + Grounding & 14.3 / 31.2 / 39.5 & 38.5 \\ \bottomrule
    \end{tabular}
\end{table}

\clearpage
\section{Language-based Visual Grounding}\label{app:visual_grounding}

\begin{minipage}{\textwidth}
    \begin{tcolorbox}[
    title=Prompt used for Visual Grounding Inference,
    boxsep=2pt,     
    left=2pt,       
    right=2pt,      
    top=2pt,        
    bottom=2pt      
    ]
        \small
        \begin{small}
\begin{verbatim}
system_prompt = (
You are a visual grounding assistant. Given an image and a question, output the 
bounding box of the image region that contains the visual information needed to 
answer the question.

The image is always 960 pixels wide and 1280 pixels tall.

Output format: [x1, y1, x2, y2] in pixels — NO other text.
  - (0, 0) is the top-left corner
  - x increases rightward (0 to 960), y increases downward (0 to 1280)
  - Make the box as tight as possible around the target
  - If no specific region applies, output the full image: [0, 0, 960, 1280]
  
Examples:
Question: 'What is the name of the store?'
Output: [95, 40, 530, 160]

Question: 'What color is the car?'
Output: [260, 580, 720, 940]

Question: 'How many windows does the building have?'
Output: [60, 180, 900, 1100]
)
\end{verbatim}
\end{small}

    \end{tcolorbox}
    \captionof{figure}{Prompt used for Visual Grounding Inference.} \label{prompt:visual_grouding}
\end{minipage}

\paragraph{Experimental Setup.} We implemented both generalized VLMs and specialized zero-shot object detection models, including Grounding-DINO and OWL-ViT. The infrastructure is built on Python 3.10 using PyTorch and the Hugging Face Transformers ecosystem. All experiments were conducted on four L40s multi-GPU cluster.

Generalized MLLMs (Llama/Qwen): To optimize throughput and memory management, these models are deployed via the \texttt{vLLM} (v0.10.1)\cite{kwon2023efficient} inference engine. We utilize a tensor-parallel configuration across available GPUs with a max model length of 8,192 tokens in bfloat16 precision. The generation process is governed by \texttt{vllm.SamplingParams} with a 75 max tokens, temperature of 0.1 and a top-p of 0.9 to promote deterministic and focused outputs.

Specialized grounding models are deployed using dedicated processors from the transformers library (\texttt{AutoProcessor} and \texttt{OwlViTProcessor}). Grounding-DINO is specifically tuned with a box threshold of 0.4 and a text threshold of 0.3 to filter low-confidence detections during the zero-shot grounding tasks. OWL-ViT uses threshold 0.1 when post processing the grounding output.

\clearpage
\section{Object Identification}
\label{app:object_identification}
\begin{minipage}{\textwidth}
    \begin{tcolorbox}[title=Prompt used for Object Identification Inference,
    boxsep=2pt,     
    left=2pt,       
    right=2pt,      
    top=2pt,        
    bottom=2pt      
    ]
        \small
        \begin{small}
\begin{verbatim}

SYSTEM_PROMPT = (
  You are a helpful visual assistant that identifies the specific target object 
  in an image referred to by a question.
  Your task is not to answer the question, but to determine which object in the 
  image the question is referring to and describe or name that object precisely.
  What is the object’s name which will help me answer the query? 
  Focus only on visual and contextual cues from the image that indicate the 
  subject of the question.
  
  Instructions:
   - Ignore the question’s semantic intent (e.g., do not explain, justify, or 
   give an opinion-based answer).
  - Identify the visual target most relevant to the question.
   - Output only the exact entity object name (e.g., 'Subaru WRX', 'The Empire 
   State Building', 'euphorbia aphylla').
   - Do not include any additional explanation, reasoning, or answer content.
   
  Example:
   Image: A photo of a blue Subaru WRX in a parking lot and the blue Subaru WRX is 
  highlighted with a green border box.
  Question: “Is this a good car for transporting seven passengers at once?”
  Correct Output: Subaru WRX
  Incorrect Output: No, the Subaru WRX can only fit 5 passengers.
   Now analyze the following image and question to output only the target object 
    name.
    
  ### Response format:
  target_object: [entity name of target object]
  
)   
\end{verbatim}
\end{small}
    \end{tcolorbox}
    \captionof{figure}{Prompt used for Object Identification Inference} \label{prompt:object_identification}
\end{minipage}

\paragraph{Experimental Setup.} We follow the same experimental setup as stage 1, visual grounding task overall (see Appendix~\ref{app:visual_grounding}). Aside of that, we further elaborate different specialized models in this stage. These specialized models utilized during the experiments are the unimodal retriever \texttt{CLIP-\allowbreak ViT-\allowbreak Large-\allowbreak Patch14-\allowbreak 336}  and the multimodal retrievers \texttt{VLM2Vec\allowbreak-V2.0} and \texttt{Qwen-3-VL-Embedding-2B}. We use the image knowledge graph\footnote{\href{https://huggingface.co/datasets/crag-mm-2025/image-search-index-public-test}{huggingface.co/datasets/crag-mm-2025/image-search-index-public-test}} from \texttt{CRAG-MM} as the retrieval corpus and recompute embeddings for each model. For all image retrievers, similary scores for query ($Q$) and candidate embeddings ($C$) are measured with cosine similarity ($\text{sim} = Q \cdot C^\top$). All retrieval indexes are built with ChromaDB\cite{chroma}.

Unimodal Retriever: For unimodal retriever, encoders only get either image or text as an input for the model. \texttt{CLIP} is under this category and we utilize image to retrieve relevant image KG information using prebuilt image index from \texttt{CRAG-MM}.

Multimodal Embedding: For unified multimodal representation for image and text question, we leverage \texttt{VLM2VEC2} and the \texttt{Qwen3-VL-Embedding-2B} models allowing for the simultaneous encoding of system instructions, text prompts, and image data into a singular embedding space. We follow experimental setups from official documents per each model. For \texttt{VLM2VECv2} ''Represent the given image with the following question'' is utilized as instruction with the regarding text input, and ''Represent the user's input.'' for the \texttt{Qwen3-VL-Embedding}

\subsection{Generalized Models (VLM)} 

\subsubsection{Object Identification and Target Saliency.}\label{app:oi_generalized}

\begin{figure}[ht!]\centering
    \includegraphics[width=0.5\linewidth]{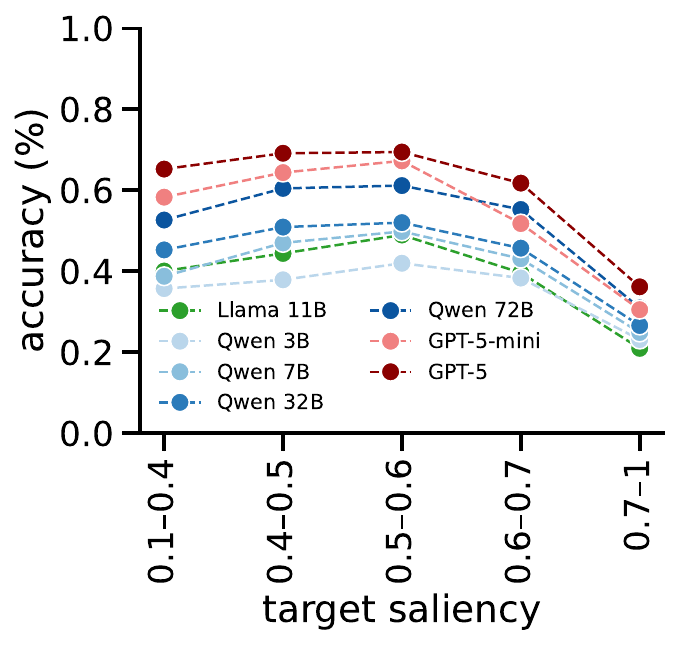}
    \caption{Generalized model (VLM) object identification accuracy with different saliency interval.}
\label{fig:oi_visual_complexity_saliency_generalized}
\end{figure}

Unlike the grounding task in Stage 1, object identification is less sensitive to visual complexity; in fact, saliency is (only weakly) negatively related to accuracy ($r_{pb}=-.168$) as shown in \cref{fig:oi_visual_complexity_saliency_generalized}. This likely stems from dataset distribution: less popular entities are more frequent in salient groups, impacting object identification more significantly. This is supported by specialized models gaining less from cropping for less popular groups compared to popular ones (\cref{fig:specialized_popularity}).

\subsection{Specialized Models (Image Retriever)}\label{app:image_retriever}

In this section, we evaluate the performance of various image retrieval models across different input modalities.
\subsubsection{Retrieval Performance across Top-$k$.}

As illustrated in \Cref{fig:imageRetriever_topK}, we compare the retrieval scalability of \texttt{CLIP-ViT-Large-Patch14-336} \cite{Radford2021LearningTV} and the multimodal retrievers \texttt{VLM2Vec-V2.0} \cite{Meng2025VLM2VecV2AM} and \texttt{Qwen-3-VL-Embedding-2B} \cite{li2026qwen3} by measuring Recall@$k$ for $k \in \{1, 5, 10, 20, 30\}$. For this comparison, all models utilize only the image input with ground-truth (GT) target region cropping to isolate the retriever's capability from potential localization errors. We observe a consistent growth in recall across all models as $k$ increases, with \texttt{Qwen-3-VL-Embedding-2B} maintaining a significant performance lead across the entire range, followed by \texttt{CLIP-ViT-Large-Patch14-336} and \texttt{VLM2Vec-V2.0}. We follow the official experimental setups for retrieval instructions when evaluating the multimodal retrievers.

\subsubsection{Sensitivity to Input Variants.} 
\Cref{tab:retrieval_results} provides a detailed breakdown of how different input modalities and preprocessing steps affect retrieval performance. Several key trends emerge:

\begin{itemize}
    \item \textit{Impact of Region Cropping:} For all models, providing a localized view of the target object results in substantial gains. Using GT crops yields the best results; for instance, boosting \texttt{Qwen-3-VL-Embedding-2B} R@1 from 22.11\% to 31.33\%. Notably, even automated cropping via G-DINO provides a performance uplift over the original full image in the CLIP baseline, indicating potential for the grounding module to further bridge the gap to GT-level performance.
    \item \textit{Multimodal vs. Unimodal Performance:} Interestingly, for the multimodal retrievers (\texttt{VLM2Vec-V2.0} and \texttt{Qwen-3-VL-Embedding-2B}), the "Image Only" input consistently outperforms the "Image + Text" combination. For example, \texttt{Qwen-3-VL-Embedding-2B} performance drops from 22.11\% to 12.62\% R@1 when the text query is added to the full image input. This suggests that the text queries may introduce noise or that current multimodal retrievers' training objectives are not yet optimized for precisely grounding the target region with textual cues to retrieve relevant information. Consequently, the visual features of the target object remain significantly more discriminative than the joint multimodal embeddings for this specific task.
    \item \textit{Text-Only Baseline:} The "Text Only" variants perform poorly across all models, with R@1 values below 2.1\%. This underscores that the dataset requires fine-grained visual perception that cannot be resolved through linguistic cues alone.
\end{itemize}


\begin{table}[t!]
\centering
\caption{Retrieval performance (Recall@$k$) comparison across different retrievers and input variants. 
"w/ G-DINO Crop" indicates that the image input is cropped based on predictions from Grounding-DINO, whereas "w/ GT Crop" denotes the use of ground-truth region of interest (ROI) annotations for cropping.
Note that CLIP is a unimodal retriever that can process either solely image or text input for retrieval, and VLM2VEC-v2.0 and Qwen3-VL-Emb. are multimodal retrievers that can process text and image together.}
\label{tab:retrieval_results}
\resizebox{\columnwidth}{!}{%
\begin{tabular}{@{} l l c c c c c @{}}
\toprule
\textbf{Model} & \textbf{Input} & \textbf{R@1} & \textbf{R@5} & \textbf{R@10} & \textbf{R@20} & \textbf{R@30} \\ \midrule
\multirow{3}{*}{CLIP} & Image & 14.97 & 26.98 & 31.85 & 38.47 & 41.25 \\
& \textit{w/ G-DINO Crop} & 17.41 & 31.77 & 36.12 & 42.91 & 45.43 \\ 
& \textit{w/ GT Crop}& \textbf{20.28} & \textbf{36.03} & \textbf{41.78} & \textbf{47.87} & \textbf{51.20} \\ \midrule
\multirow{5}{*}{VLM2VEC-v2.0} & 
Image & 12.36 & 22.72 & 26.72 & 32.03 & 35.25 \\
& \textit{w/ GT Crop} & \textbf{15.75} & \textbf{29.24} & \textbf{34.38} & \textbf{40.03} & \textbf{43.60} \\
& Text & 1.65 & 4.00 & 6.01 & 9.57 & 11.92 \\ 
& Image + Text & 8.88 & 17.06 & 20.54 & 26.20 & 28.20 \\
& \textit{w/ Image (GT Crop) + Text} & 9.40 & 19.50 & 24.11 & 29.24 & 32.29 \\
\midrule
\multirow{5}{*}{Qwen3-VL-Emb} 
& Image & 22.11 & 36.99 & 42.99 & 49.35 & 51.61 \\
& \textit{w/ GT Crop} & \textbf{31.33} & \textbf{46.30} & \textbf{51.87} & \textbf{56.92} & \textbf{58.83} \\
& Text & 2.09 & 5.48 & 7.75 & 10.97 & 13.40 \\ 
& Image + Text & 12.62 & 23.24 & 29.77 & 36.99 & 40.21 \\
& \textit{w/ Image (GT Crop) + Text} & 18.54 & 30.64 & 36.73 & 42.12 & 45.52 \\
\bottomrule
\end{tabular}%
}
\end{table}

\subsubsection{Effect of Popularity.}
\begin{figure}[t]
    \centering
    \begin{minipage}{0.48\linewidth}
        \centering
            \includegraphics[width=\linewidth]{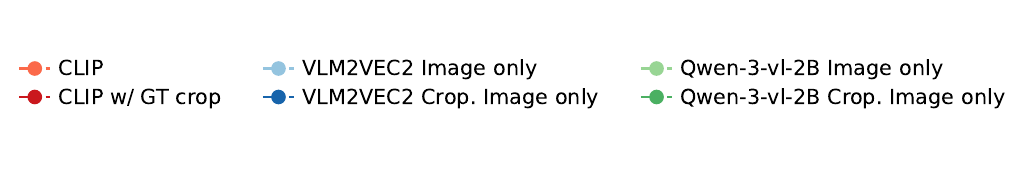}
        \includegraphics[width=\linewidth]{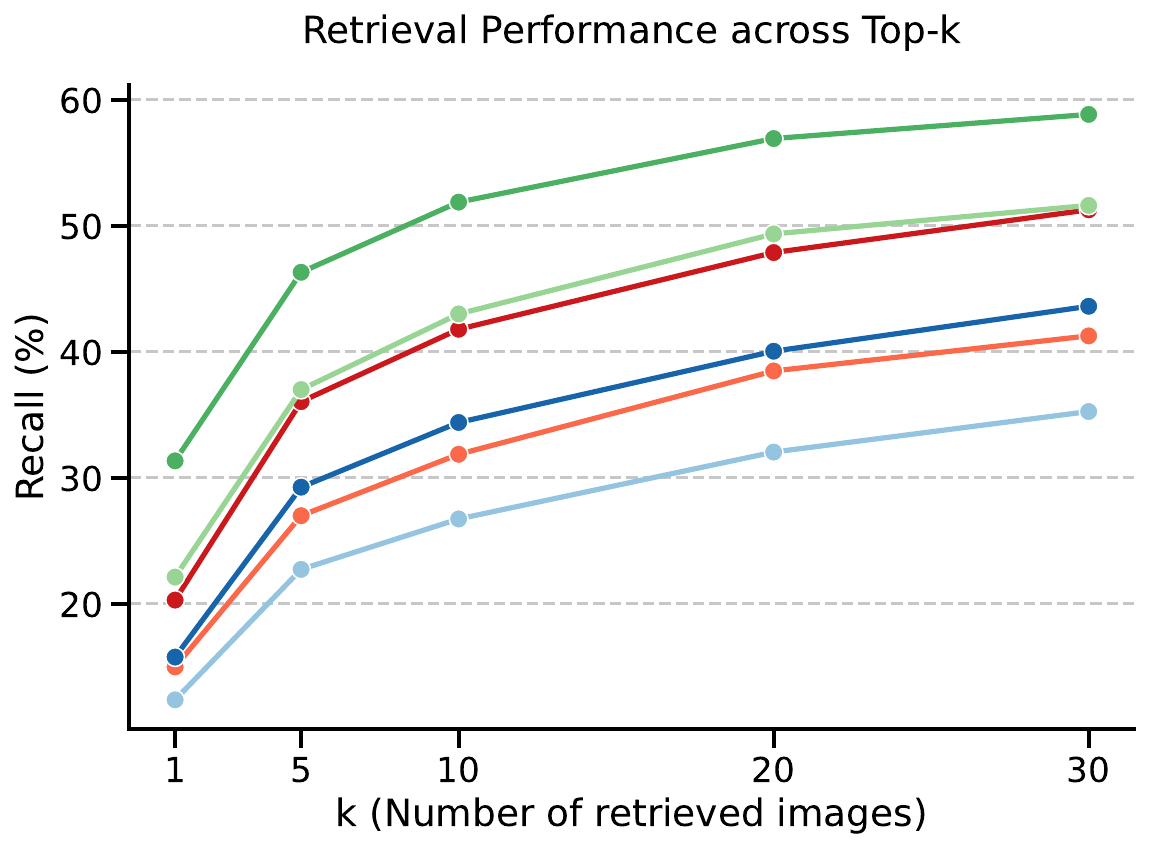}
        \caption{Image retriever results with different top k. For comparison, all retrievers utilize image only input and ground truth target region cropped version.}
        \label{fig:imageRetriever_topK}
    \end{minipage}
    \hfill
    \begin{minipage}{0.48\linewidth}
        \centering
        \includegraphics[width=\linewidth]{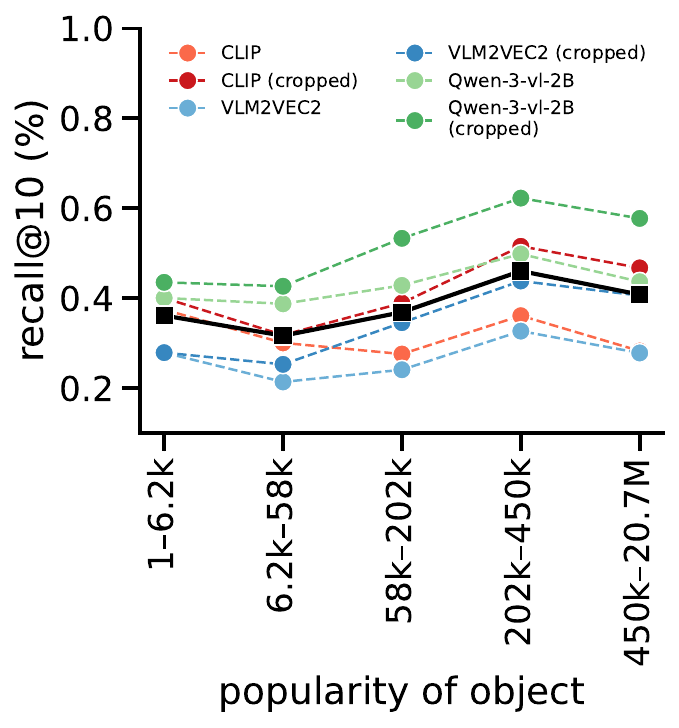}
        \caption{Specialized image retriever object identification results with different entity popularity interval. For comparison, all retrievers utilize image only input and ground truth target region cropped version. Black rectangle means model mean.}
        \label{fig:specialized_popularity}
    \end{minipage}
\end{figure}

\Cref{fig:specialized_popularity} illustrates specialized image retrieval performance trends stratified by object popularity.


\clearpage
\section{Knowledge Retrieval and Reasoning}
\begin{minipage}{\textwidth}
    \begin{tcolorbox}[title=Prompt used for Knowledge Extraction Inference,
    boxsep=2pt,     
    left=2pt,       
    right=2pt,      
    top=2pt,        
    bottom=2pt      
    ]
        \small
        \begin{small}
\begin{verbatim}
SYSTEM_PROMPT = (
  You are a helpful assistant that truthfully answers user questions. Keep your 
  response concise and to the point.
)
\end{verbatim}
\end{small}

    \end{tcolorbox}
    \captionof{figure}{Prompt used for Knowledge Extraction Inference} \label{prompt:knowledge_extraction}
\end{minipage}

\paragraph{Experimental Setup.} We maintain an experimental configuration consistent with the stage \circled{1} visual grounding and stage \circled{2} object identification tasks (see \Cref{app:visual_grounding,app:object_identification}). For the additional text retrieval component, we employ \texttt{BGE\allowbreak-large\allowbreak-en\allowbreak-v1.5} as the primary text encoder. We leverage the prebuilt Web search index provided by \texttt{CRAG-MM}\footnote{\href{https://huggingface.co/datasets/crag-mm-2025/web-search-index-public-test}{huggingface.co/datasets/crag-mm-2025/web-search-index-public-test}}, which serves as the candidate Web corpus for our stage \circled{3} knowledge retrieval and reasoning experiments.
\section{Evaluation} \label{app: evaluation}
Figure~\ref{prompt:llm-as-a-judge} shows the prompt used for LLM-as-a-judge evaluation.

\begin{figure}[t]
\centering
\begin{tcolorbox}[
  title=Prompt used for LLM-as-a-judge evaluation,
  width=\textwidth,
  boxsep=2pt,     
    left=2pt,       
    right=2pt,      
    top=2pt,        
    bottom=2pt      
]
\footnotesize
\input{Appendix/evaluation_prompt}
\end{tcolorbox}
\caption{Prompt used for LLM-as-a-judge}
\label{prompt:llm-as-a-judge}
\end{figure}
\clearpage
\section{Manual Error Analysis}\label{app:error_analysis}

To investigate the systemic failure modes of bimodal and grounded RAG pipelines, even when utilizing ground-truth (GT) region cropping prior to image retrieval, we conduct a qualitative manual analysis across two distinct backbones: \texttt{GPT-5} and \texttt{Qwen-2.5-VL-32B}. We randomly sample 200 failure cases identified from the test instances. This evaluation allows us to categorize recurring bottlenecks across multi-stage subtasks, including grounding, object identification, and knowledge extraction. While both models heavily suffer from knowledge-retrieval bottlenecks, distinct architectural trends emerge: \texttt{Qwen-2.5-VL} is primarily constrained by fine-grained object identification, whereas \texttt{GPT-5} failures are significantly harder to isolate, where full statistical distributions are summarized in \Cref{tab:error_case_categorization_two_models}.

The primary failure dimensions are detailed below, with representative qualitative examples provided in \Cref{tab:error_case_examples}:

\begin{description}
    \item[Difficult Attribution:] 
    These cases involve errors where the failure cannot be clearly assigned to a single component. For example, when asked about the number of generations for a \textit{Nissan Armada}, the model predicted just ``Five'' despite the ground truth being three. The difficulty in diagnosing whether the model failed during object identification or knowledge retrieval underscores our motivation for decoupling KI-VQA evaluation, as aggregate metrics often obscure whether a failure stems from retrieval or internal model bias.

    \item[Knowledge Bottlenecks:] These errors occur when the model correctly identifies the target object but fails to retrieve or utilize specific, fine-grained factual data. In Table~\ref{tab:error_case_examples}, the \textit{Toyota} example illustrates this: while the model identifies the vehicle correctly, it fails to accurately retrieve the founder's birth date, providing a hallucinated response instead. This highlights that even state-of-the-art models struggle with the `long tail' of domain-specific knowledge without more precise retrieval-augmentation.

    \item[Object Identification Failures:] Identification errors are characterized by an inability to distinguish between similar categories or recognize non-salient targets. Notably, despite the localized focus provided by GT-region cropping, both specialized retrievers and the generalized VLM occasionally failed to correctly identify the primary subject. This suggests a critical need for more robust, universal object recognizers capable of handling fine-grained taxonomies.
    \begin{itemize}
        \item \textbf{Fine-grained confusion:} As seen in the \textit{Panther \allowbreak chameleon} case, the model misidentified the subject as a \textit{Jackson's \allowbreak chameleon}.
        \item \textbf{Saliency issues:} For the \textit{Flag of Missouri}, the model failed to recognize the specific flag in a complex outdoor scene, defaulting to a more common \textit{French flag}.
    \end{itemize}

    \item[Reasoning and Ambiguity:] Unlike knowledge bottlenecks, these failures occur when the model possesses the correct information but fails to process it logically, or when the query is fundamentally unclear.
    \begin{itemize}
        \item \textbf{Incomplete Reasoning:} In the \textit{Masada} example, the model identifies the builder correctly as ``Herod the Great'' but fails to provide the nuanced explanation required by the context, even when the retriever provides broader details (e.g., citing both Alexander Jannaeus and Herod). Note that retrieval correctly retrieve sufficient information \textit{(Masda: "builder": "[[Alexander Jannaeus]] [[Herod the Great]]",)}, but fails to augment these information while answering.
        \item \textbf{Inherent Ambiguity:} The \textit{Paper towel} case demonstrates a failure where the model provides a technical material breakdown (wood pulp, cardboard) but fails to synthesize a direct, concise answer to the user's intent caused from ambiguity of the nature of referring expression ``this'' in this case.
    \end{itemize}

\end{description}

\begin{table}[ht!]
\centering
\caption{Error analysis of grounded bimodal RAG pipeline.}
\label{tab:error_case_categorization_two_models}
\resizebox{\columnwidth}{!}{%
\begin{tabular}{lcc}
\toprule
\textbf{Failure Category} & \textbf{GPT-5 (\%)} & \textbf{Qwen-2.5-VL-32B(\%)} \\
\midrule
Difficult Attribution & 41 & 1 \\
Knowledge Bottlenecks & 21 & 23 \\
Object Identification Failures: Fine-grained confusion & 20 & 45 \\
Incomplete Reasoning & 6 & 14 \\
Object Identification Failures: Saliency issues & 5 & 10 \\
Inherent Ambiguity & 3 & 6 \\
Wrong Judge & 4 & 1 \\
\bottomrule
\end{tabular}%
}
\end{table}

\begin{table}[ht!]
\centering
\caption{Qualitative analysis of failure cases in the Bimodal and grounded RAG pipeline (GPT-5 backbone, GT region cropping). Categorization based on manual error case analysis.}
\label{tab:error_case_examples}
\setlength{\tabcolsep}{8pt}
\resizebox{1.0\linewidth}{!}{
\begin{tabular}{m{3.3cm}|m{5cm}m{5.5cm}m{5.5cm}m{5.5cm}}
        \toprule
         \multicolumn{1}{c|}{Category} & \multicolumn{1}{c}{Image} &\multicolumn{1}{c}{Target Object} &\multicolumn{1}{c}{Question \& Answer}&\multicolumn{1}{c}{Prediction} \\ 
        \midrule
        \multicolumn{1}{c|}{Difficult Attribution} &  \centering
        \includegraphics[width=45mm,height=45mm]{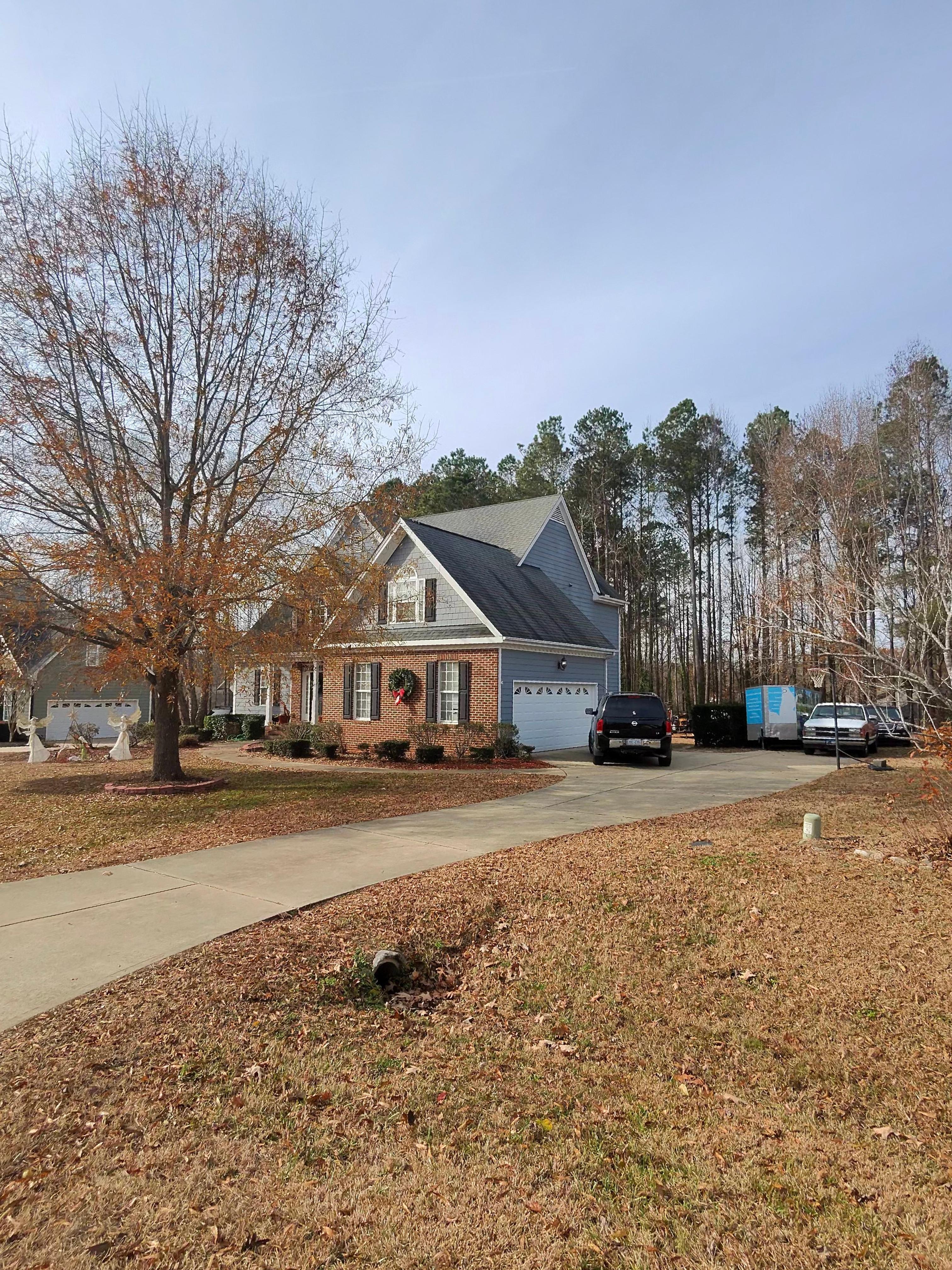} & \centering Nissan Armada \href{https://en.wikipedia.org/wiki/Nissan_Armada}{en.wikipedia.org/wiki/Nissan\_Armada}& \textbf{Q:} How many different generations of this car are there?
\newline \textbf{A:} As of 2025, there have been three generations of the Nissan Armada, with the first generation entering production in 2003, the second in 2016, and the third being in production since 2024.&
        Five. \tabularnewline
        
        \midrule 
        \multicolumn{1}{c|}{%
          \begin{tabular}{@{}c@{}} Object ID Failure \\ (Fine-grained Confusion) \end{tabular}%
        } &
        \centering
        \includegraphics[width=45mm,height=45mm]{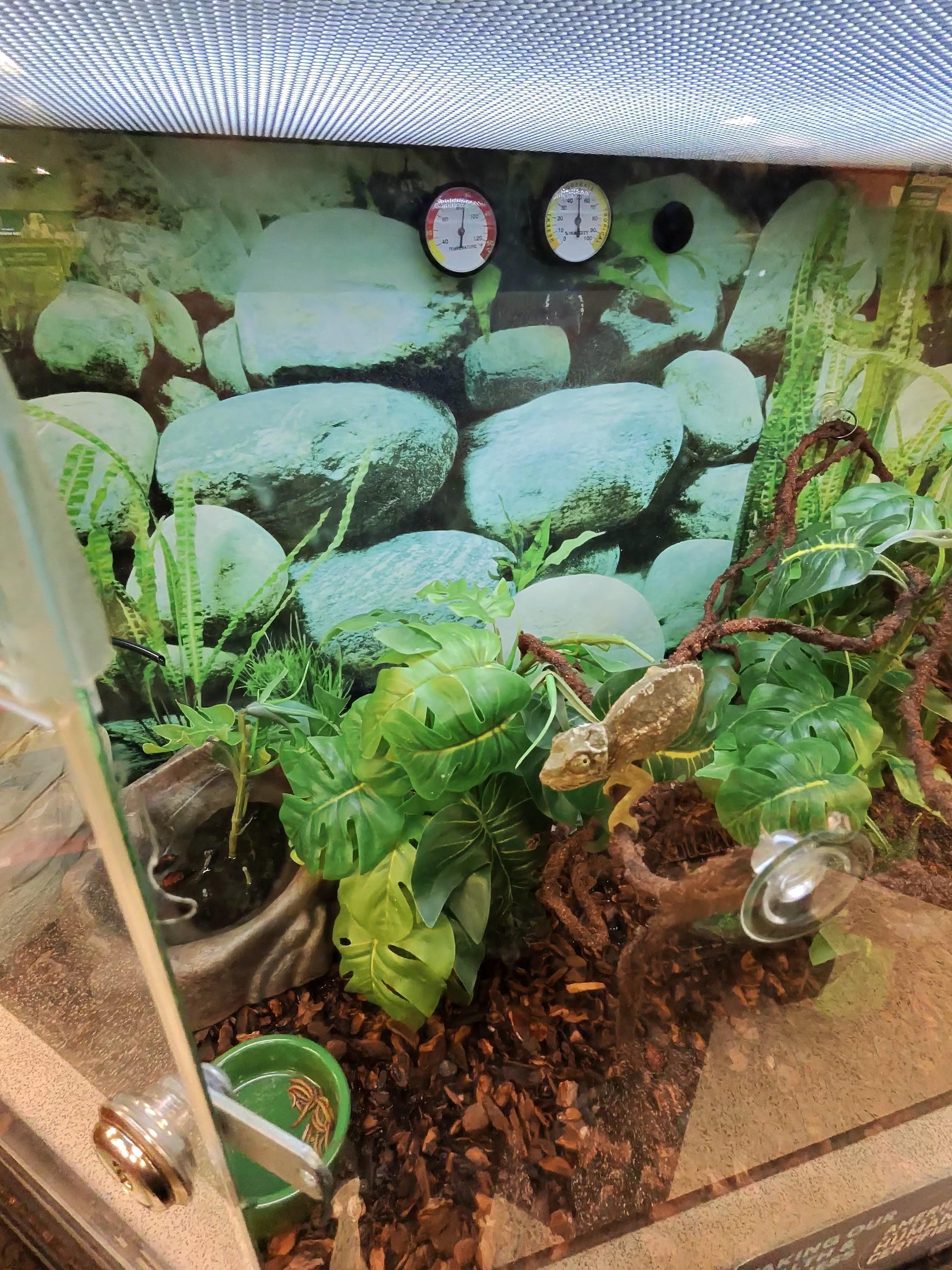} & 
        \centering Panther chameleon \href{https://en.wikipedia.org/wiki/Panther_chameleon}{en.wikipedia.org/wiki/Panther\_chameleon}& 
        \textbf{Q:} what is a difference between this and a veiled chameleon? \newline \textbf{A:} the panther chameleon and the veiled chameleon have different dispositions with the panther chameleon being more friendly.&
        It looks like a Jackson’s chameleon. Key differences from a veiled chameleon: ... 
        \tabularnewline
        
        \midrule
        \multicolumn{1}{c|}{Knowledge Bottleneck} &  \centering
        \includegraphics[width=45mm,height=45mm]{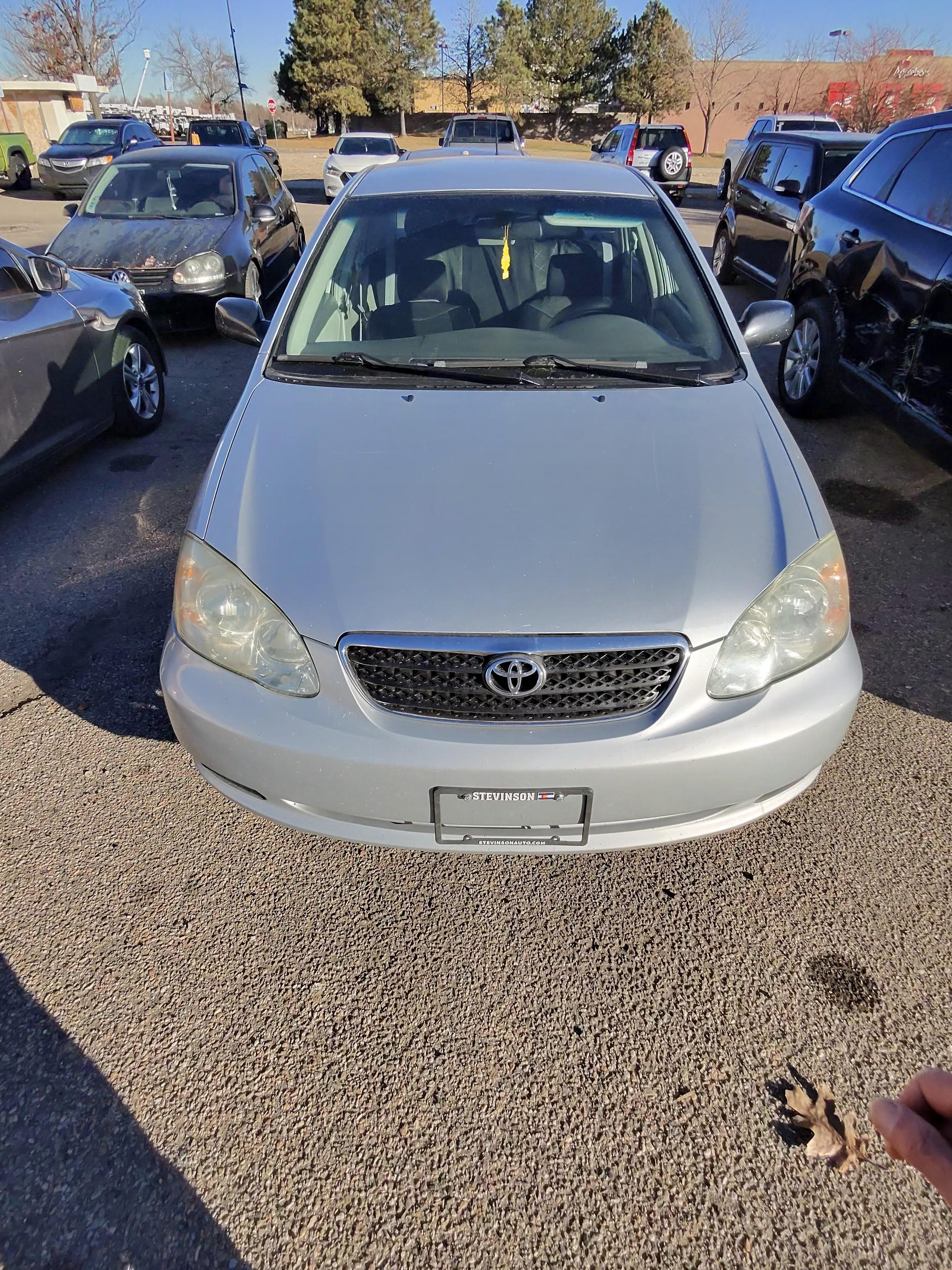} & 
        \centering Toyota\\
        \href{https://en.wikipedia.org/wiki/Toyota}{en.wikipedia.org/wiki/Toyota}& 
        \textbf{Q:} when was the founder of this company born? \newline \textbf{A:} sakichi toyoda was born march 19, 1867.&
        That’s a Toyota. Toyota Motor Corporation’s founder, Kiichiro Toyoda, was born on June 11, 1894. (Sakichi Toyoda, founder of the Toyota Group, was born Feb 14, 1867.)
        \tabularnewline
        
        \midrule
        \multicolumn{1}{c|}{%
          \begin{tabular}{@{}c@{}} Object ID Failure \\ (Non-salient) \end{tabular}%
        } &
        \centering
        \includegraphics[width=45mm,height=45mm]{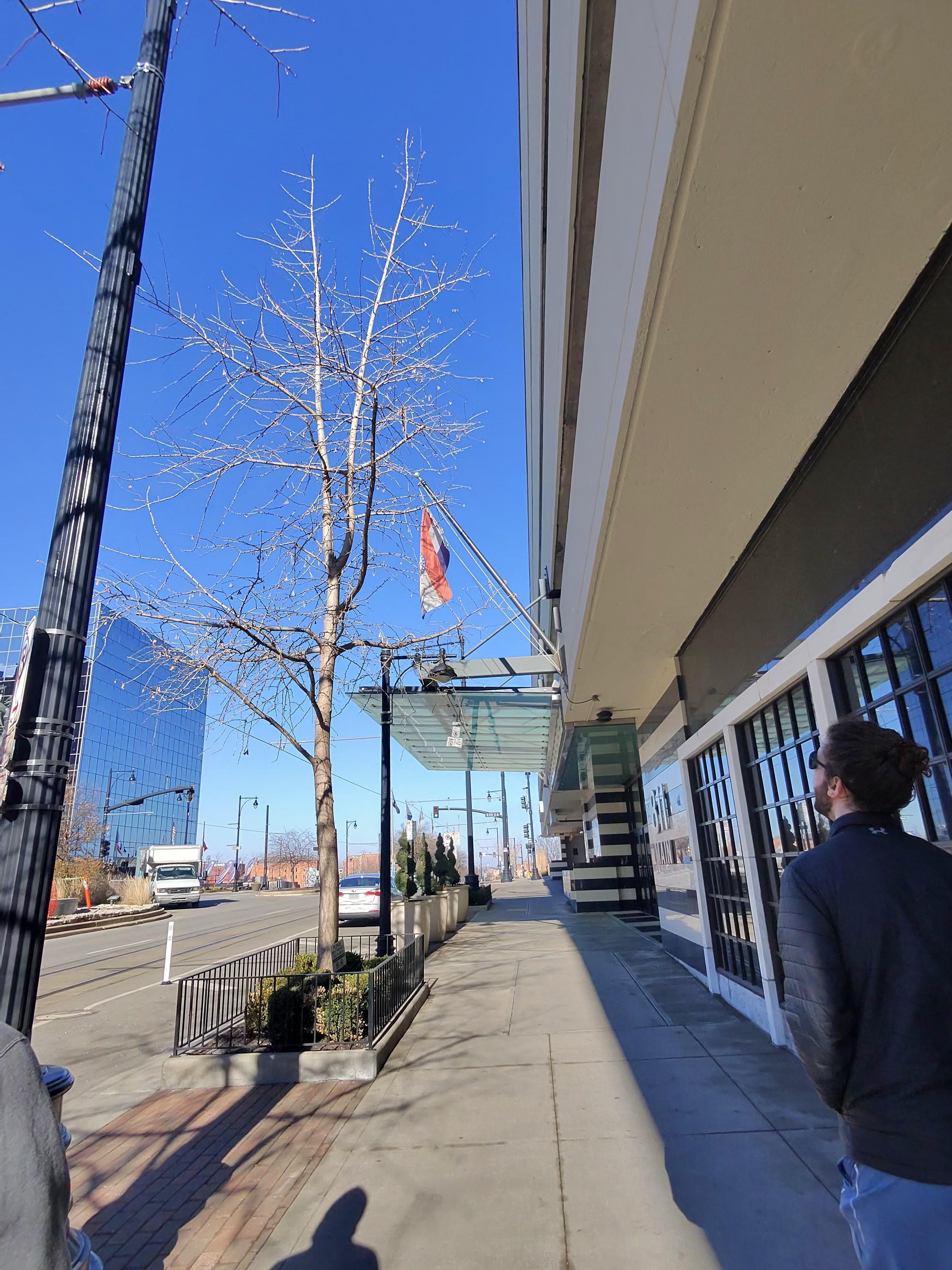} & 
        \centering Flag of Missouri \href{https://en.wikipedia.org/wiki/Flag_of_Missouri}{en.wikipedia.org/wiki/Flag\_of\_Missouri}& 
        \textbf{Q:} when was this flag created? \newline \textbf{A:} the flag of missouri was adopted in 1913.&
        It looks like the French flag. The modern tricolor design was officially adopted on February 15, 1794.
        \tabularnewline
        
        \midrule
        \multicolumn{1}{c|}{Incomplete Reasoning} &  \centering
        \includegraphics[width=45mm,height=45mm]{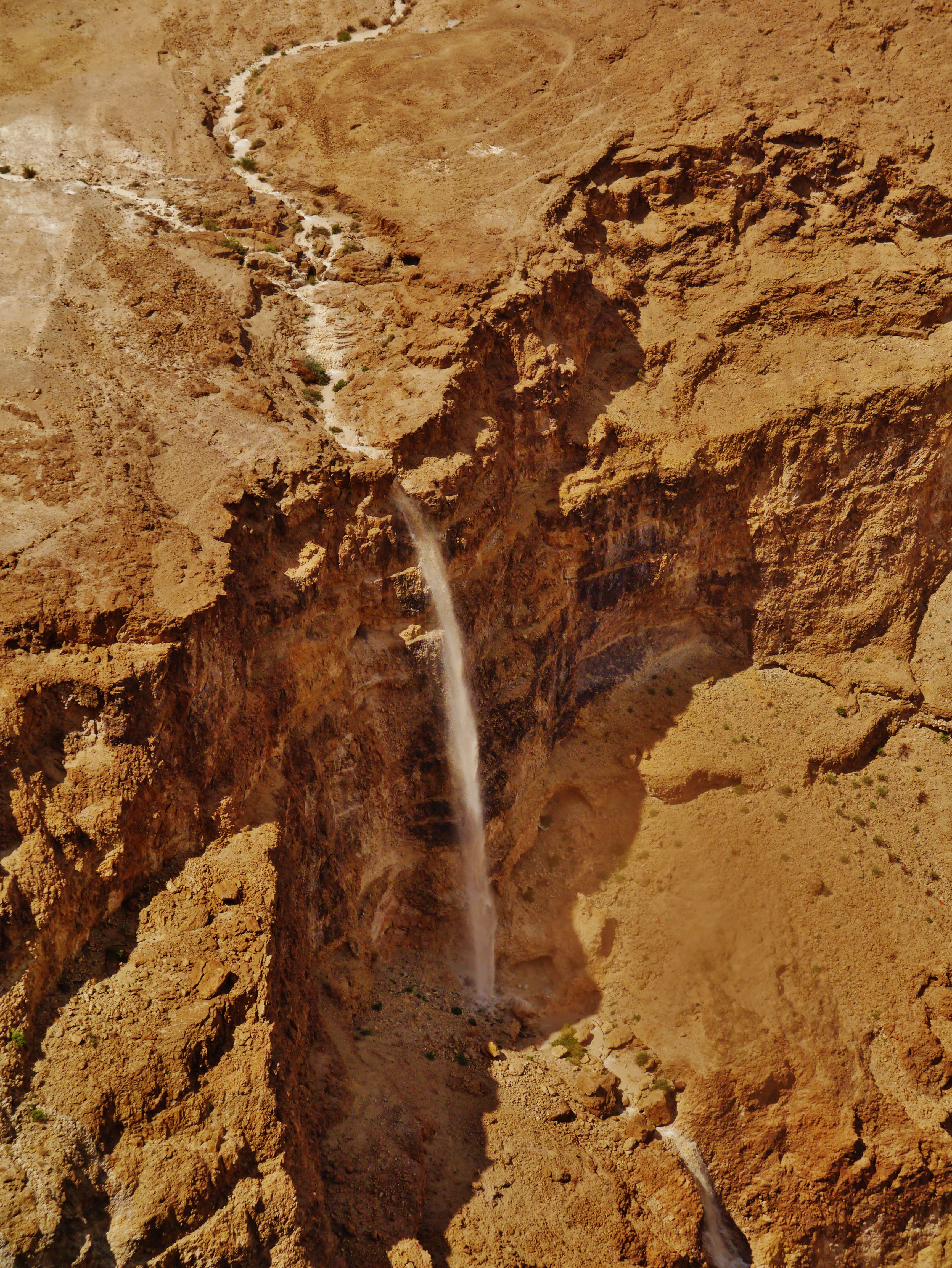} & 
        \centering Masada \href{https://en.wikipedia.org/wiki/Masada}{en.wikipedia.org/wiki/Masada}& 
        \textbf{Q:} who is credited with building this ancient fortress? \newline \textbf{A:} masada was built by herod the great, although some sources suggest it may have been initially constructed by alexander jannaeus.&
        Herod the Great.
        \tabularnewline

        \midrule
        \multicolumn{1}{c|}{Ambiguity} &  
        \centering \includegraphics[width=45mm,height=45mm]{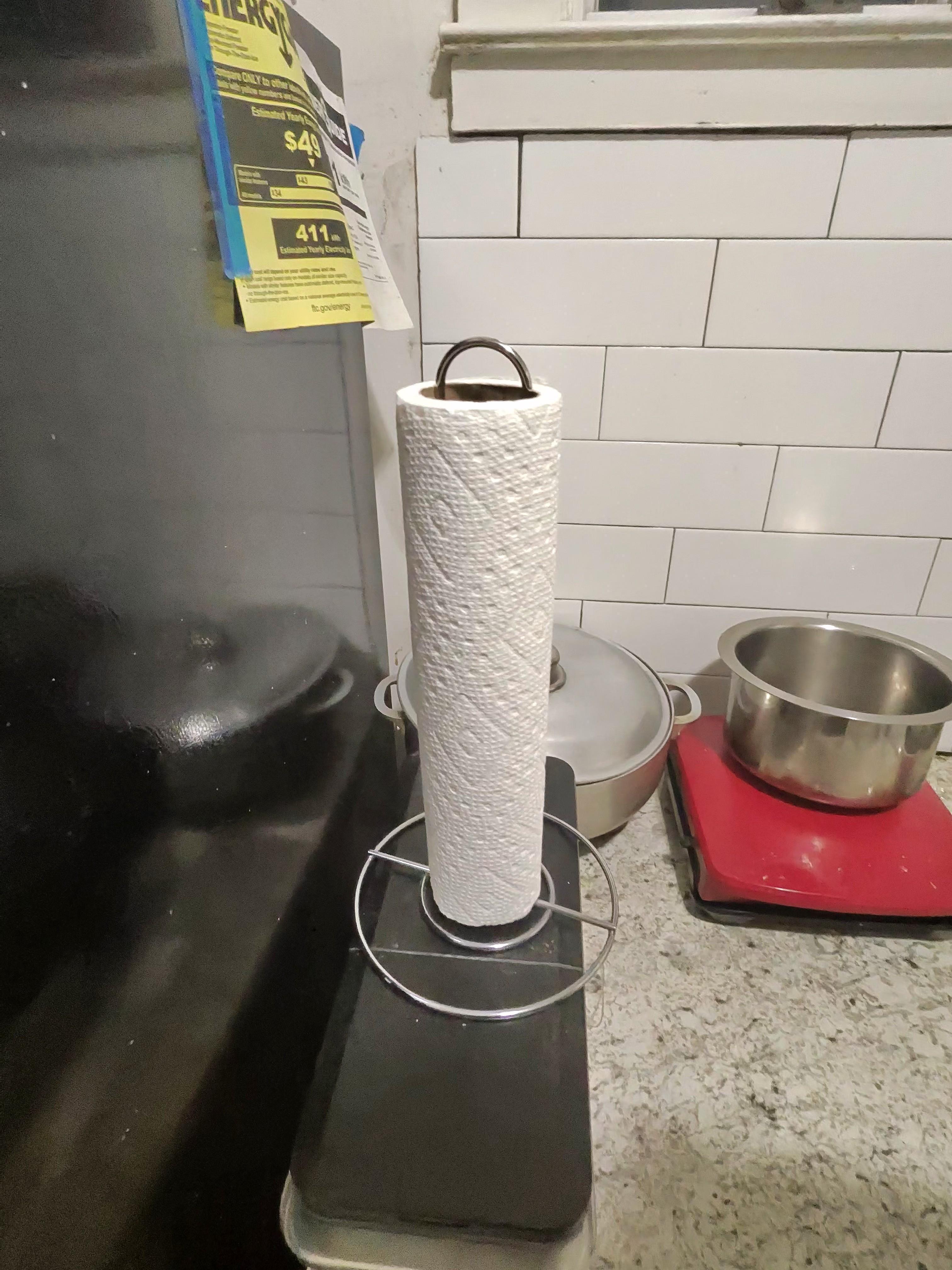} &
        \centering Paper towel \href{https://en.wikipedia.org/wiki/Paper_towel}{en.wikipedia.org/wiki/Paper\_towel} &
        \textbf{Q:} what is this made of? \newline \textbf{A:} paper towels are made from either virgin or recycled paper pulp which is extracted from wood or fiber crops. they are sometimes bleached during the production process to lighten coloration, and may also be decorated with colored images on each square.&
        - The roll is paper towels, made from paper (wood pulp).
- The inner tube is cardboard.
- The stand/holder appears to be metal wire (likely stainless or chrome-plated steel).
        \tabularnewline
        \bottomrule
\end{tabular}
}
\end{table}

\end{document}